\documentclass[10pt,journal,compsoc]{IEEEtran}
\ifCLASSOPTIONcompsoc
  \usepackage[nocompress]{cite}
\else
  \usepackage{cite}
\fi
\ifCLASSINFOpdf
\else
\fi
\usepackage{bbding}

\usepackage{amsfonts}

\usepackage{amsmath}

\usepackage{graphicx}

\usepackage{float}  

\usepackage{subfigure}

\usepackage{bm}

\usepackage{courier}

\usepackage{booktabs}

\usepackage{multirow}

\usepackage{rotating}

\usepackage{amsthm}

\usepackage{footnote}

\usepackage{color}

\usepackage{stfloats}

\usepackage{threeparttable}

\usepackage{adjustbox}

\usepackage{amssymb}

\usepackage[table]{xcolor}

\newtheorem{proposition}{Proposition}

\newtheorem{definition}{Definition}

\newtheorem{property}{Property}

\newtheorem{example}{Example}

\newtheorem{theorem}{Theorem}

\newtheorem{lemma}{Lemma}

\usepackage{color}

\definecolor{mygray}{gray}{.85}
\definecolor{mygray1}{gray}{.7}
\definecolor{mygray2}{gray}{.93}

\newcommand{\tabincell}[2]{\begin{tabular}{@{}#1@{}}#2\end{tabular}}

\makeatletter
\newcommand{\thickhline}{%
	\noalign {\ifnum 0=`}\fi \hrule height 1pt
	\futurelet \reserved@a \@xhline
}
\makeatother

\begin{document}
%
\title{Representing Noisy Image Without Denoising}
%
%
%
%

\author{Shuren~Qi,~Yushu~Zhang,~\IEEEmembership{Senior~Member,~IEEE},~Chao~Wang,~Tao~Xiang,~\IEEEmembership{Senior~Member,~IEEE},\\~Xiaochun~Cao,~\IEEEmembership{Senior~Member,~IEEE},~and~Yong~Xiang,~\IEEEmembership{Senior~Member,~IEEE}%
\IEEEcompsocitemizethanks{
	\IEEEcompsocthanksitem S. Qi and Y. Zhang are with the College of Information Technology, Jiangxi University of Finance and Economics, Nanchang 330013, China, and also with the College of Computer Science and Technology, Nanjing University of Aeronautics and Astronautics, Nanjing 210016, China (e-mail: {shurenqi, yushu}@nuaa.edu.cn).
	\IEEEcompsocthanksitem C. Wang is with the College of Computer Science and Technology, Nanjing University of Aeronautics and Astronautics, Nanjing 210016, China (e-mail: c.wang@nuaa.edu.cn).
	\IEEEcompsocthanksitem T. Xiang is with the College of Computer Science, Chongqing University, Chongqing 400044, China (e-mail: txiang@cqu.edu.cn).
	\IEEEcompsocthanksitem X. Cao is with the School of Cyber Science and Technology, Shenzhen Campus of Sun Yat-sen University, Shenzhen 510275, China (e-mail: caoxiaochun@mail.sysu.edu.cn).
	\IEEEcompsocthanksitem Y. Xiang is with the Deakin Blockchain Innovation Laboratory, School of Information Technology, Deakin University, Burwood, VIC 3125, Australia (e-mail: yong.xiang@deakin.edu.au).}

	\thanks{This work was supported by the National Natural Science Foundation of China under Grants 62025604, 62072062, and U20A20176.}
	\thanks{(Corresponding author: Yushu Zhang.)}
}

%
%

\markboth{S. Qi \MakeLowercase{\textit{et al.}}: Representing Noisy Image Without Denoising} {S. Qi \MakeLowercase{\textit{et al.}}: Representing Noisy Image Without Denoising}
%



\IEEEtitleabstractindextext{%
\begin{abstract}
A long-standing topic in artificial intelligence is the effective recognition of patterns from noisy images. In this regard, the recent data-driven paradigm considers 1) improving the representation robustness by adding noisy samples in training phase (i.e., data augmentation) or 2) pre-processing the noisy image by learning to solve the inverse problem (i.e., image denoising). However, such methods generally exhibit inefficient process and unstable result, limiting their practical applications. In this paper, we explore a non-learning paradigm that aims to derive robust representation directly from noisy images, without the denoising as pre-processing. Here, the noise-robust representation is designed as Fractional-order Moments in Radon space (FMR), with also beneficial properties of orthogonality and rotation invariance. Unlike earlier integer-order methods, our work is a more generic design taking such classical methods as special cases, and the introduced fractional-order parameter offers time-frequency analysis capability that is not available in classical methods. Formally, both implicit and explicit paths for constructing the FMR are discussed in detail. Extensive simulation experiments and robust visual applications are provided to demonstrate the uniqueness and usefulness of our FMR, especially for noise robustness, rotation invariance, and time-frequency discriminability.
\end{abstract}

}

\maketitle

\IEEEdisplaynontitleabstractindextext

%
\IEEEpeerreviewmaketitle

\IEEEraisesectionheading{\section{Introduction}\label{sec:intro}}

%
%
%
%

\IEEEPARstart{H}{andling} degraded image versions of the original scene is a fundamental and challenging requirement in numerous computer visual tasks \cite{ref1}. One major class of degradation models is image noise, where observed noisy image ${f_\eta }$ is the result of ideal scene $f$ degraded by a noise $\eta$: $f_\eta = f + \eta $. 

In a modern perspective, image noise mainly occurs at three phases over the life cycle: 1) \emph{photographic noise} arises from the imperfect imaging, e.g., a fast shutter speed in low illumination environment \cite{ref2}; 2) \emph{transmission noise} is introduced over the channel, e.g., image pre-processing in online social media \cite{ref3}; and 3) \emph{editing noise} is an active artificial perturbation, which may be introduced by an adversary with editing software, e.g. Photoshop, for misleading system decisions \cite{ref4}. Therefore, artificial vision systems are expected to be robust against image noise, motivated by the principles of effectiveness and security \cite{ref5}.

\subsection{State of the Art and Motivation}

Robust visual systems heavily rely on robust data representations \cite{ref6, ref7}. For an image representation with noise robustness, current data-driven methods, mostly Convolutional Neural Networks (CNN), often resort to the solutions of \emph{data augmentation} and \emph{denoising pre-processing} \cite{ref8}.

In the training phase, the data augmentation strategy directly adds noisy image samples to the training set, prompting the CNN to learn noise patterns. This method is somewhat effective when the training and testing data are with generally consistent noise patterns; conversely, for unseen noise patterns, the noise robustness will no longer hold \cite{ref9}. It is also clear that such augmentation significantly increases the training cost, while lacking guarantees on explainability. In the testing phase, the denoising-based image restoration, i.e., estimating $\hat f$ from observation ${f_\eta }$ to approximate ideal scene $f$, can serve as a pre-processing of the CNN representation. Such denoising module contributes to the explainability, and also being directly compatible with existing representations without retraining. However, due to the \emph{ill-posed} nature of the inverse problem \cite{ref10, ref11}, the denoising results are typically unstable and introduce new errors (to which the CNN is also sensitive) \cite{ref12, ref13}. Therefore, in practice, the denoising pre-processing strategy is unable to provide a satisfactory robustness against image noise. As for the efficiency, the additional inference cost for each sample is also not negligible.

Another paradigm aims to derive robust representation directly from degraded image versions, without the brute-force learning or restoration as pre-processing.

The idea of \emph{in-form} robustness can be traced back to the 1960’s, when Hu \cite{ref14} introduced the seven geometric invariants from statistical moments. After 60 years of research, \emph{moments and moment invariants} have become the classical representation strategy for small-scale robust vision problems \cite{ref15}. In this field, early efforts, including Hu’s invariants, achieved invariance to multiple geometric transformation groups, but were very sensitive to image noise due to the non-orthogonality of the basis \cite{ref16}. For this light, the community has developed a series of orthogonal moments and moment invariants, including Zernike Moments (ZM) \cite{ref16}, Orthogonal Fourier-Mellin Moments (OFMM) \cite{ref17}, Chebyshev-Fourier Moments (CHFM) \cite{ref18}, Jacobi-Fourier Moments (JFM) \cite{ref19}, Exponent-Fourier Moments (EFM) \cite{ref20}, Polar Complex Exponential Transform (PCET) \cite{ref21}, and Bessel-Fourier Moments (BFM) \cite{ref22}. In real-world applications of invariant pattern recognition, these orthogonal representations are well known to have a common beneficial property, i.e., geometric invariance along with certain noise robustness. From a Fourier perspective, however, such robustness is restricted, where only a few low-order moments capturing low-frequency information are stable while the rest remain sensitive to noise \cite{ref23}. When more moments are needed for a more informative representation, one must to face a tricky contradiction between discriminability and noise robustness. To alleviate this contradiction, orthogonal moments and moment invariants can be extended from the image domain to the Radon domain, where the latter is with inherently higher Signal-to-Noise Ratio (SNR) \cite{ref24}. Such the \emph{Moments in Radon} (MR) have been verified to have an improved noise robustness, while maintaining the advantages of classical moments in image domain, e.g., orthogonality and geometric invariance.

\emph{Motivation.} To the best of our knowledge, mathematically, the above researches of MR generally focus on integer-order cases only. As for a practical perspective, this property will further lead to a limited flexibility and discriminability for the MR-based representations, especially the spatial information is neglected.

\begin{table*}[t]
	\caption{Theoretical Comparison on Beneficial Properties With Related Methods.}
	\centering
	\begin{tabular}{cccccc}
		\thickhline
		Method & \tabincell{c}{Radon-domain\\ simple statistics\\ w/o form (6) \\(e.g. \cite{ref28, ref29})} & \tabincell{c}{Radon-domain\\ orthogonal transforms\\ w/ form (6) \\(e.g. \cite{ref30, ref31, ref32})} & \tabincell{c}{Image-domain\\ fractional-order\\ moments w/ form (6) \\(e.g. \cite{ref23, ref35})} & \tabincell{c}{Radon-domain\\ integer-order\\ moments w/ form (6) \\(e.g. \cite{ref24})} & \tabincell{c}{Radon-domain\\ fractional-order\\ moments w/ form (6) \\(i.e. our work)} \\
		\midrule
		\rowcolor{mygray2}\tabincell{c}{Generic\\ nature} &   &   & \checkmark     &   & \checkmark \\
		\tabincell{c}{Rotation\\ invariance} &    & \checkmark     & \checkmark     & \checkmark     & \checkmark \\
		\rowcolor{mygray2}\tabincell{c}{Robustness\\ to noise} & \checkmark     & \checkmark     &   & \checkmark     & \checkmark \\
		\tabincell{c}{Time-frequency\\ discriminability} &  &  & \checkmark     &     & \checkmark \\
		\bottomrule
	\end{tabular}%
\end{table*}

\subsection{Contributions}

Motivated by above facts, we attempt to present a more comprehensive study on the image robust representation for challenging noise corruptions. In this paper, a new time-frequency discriminative image representation is proposed with in-form noise robustness and geometric invariance, eliminating the need for any learning or denoising operations. It has potential applications in various small-scale robust vision problems, especially security and forensic applications with adversary assumptions.

In general, our work can be considered as a theoretical and practical extension for the state-of-the-art MR research.

\emph{Theory.} We generalize the approach of MR from integer-order space to fractional-order space for the first time, formulating the Fractional-order Moments in Radon (FMR) implicitly and explicitly. Note that this FMR is a more generic design treating the MR methods as special cases, and also maintaining the beneficial noise robustness and geometric invariance; more importantly, the introduced fractional-order parameter in FMR offers a time-frequency discriminability that is not available in MR methods.

\emph{Practice.} We comprehensively benchmark the proposed FMR and its image-domain counterpart, as well as lightweight and deep learning representations (with augmentation/denoising), on a pattern recognition task under challenging noise corruptions. Note that such overall robustness statistics, especially covering learning representations, have not been given until this work. Furthermore, we also introduce the FMR representation in robust visual tasks, i.e., template matching in the wild and zero-watermarking for copyright protection. Here, both benchmark experiments and real-world applications confirm the effectiveness of FMR w.r.t. noise robustness, rotation invariance, and time-frequency discriminability.

\section{Related Works}

We briefly review the closely related Radon-domain representation and fractional-order orthogonal moments, along with image denoising as another paradigm and robust visual tasks as downstream applications.

\subsection{Radon-domain Representation}

The Radon transform \cite{ref25} is a classical image projection, and is widely used in the reconstruction problem of medicine (medical imaging), engineering (electron microscopy), physical science (optics), and material science (crystallography). In computer vision and pattern recognition, the Radon transform has also received attention for its orthogonality, low complexity, noise robustness, and beneficial properties concerning geometric transformations.

Initially, some Radon-domain representations, e.g., $R$-signature \cite{ref26}, $\Phi $-signature \cite{ref27}, RCF \cite{ref28}, and HRT \cite{ref29}, are built from simple statistics, e.g., summation \cite{ref26, ref27}, maximum \cite{ref28}, or histogram \cite{ref29}, of the Radon coefficients. In general, such non-complete methods may discard discriminative information of the Radon, leading to limitations for informative vision tasks. For this reason, other Radon-domain representations, e.g., R2DFM \cite{ref30}, RWF \cite{ref31}, RMF \cite{ref32}, and MR \cite{ref24}, resort to complete and orthogonal transforms, e.g., Fourier transform \cite{ref30}, wavelet transform \cite{ref31}, Mellin transform \cite{ref32}, or orthogonal moments \cite{ref24}. Here, the MR framework covers a variety of potential basis functions, and it can be considered as a generalization of R2DFM and RMF w.r.t. representation definition and invariance analysis. 

Note that such representations based on orthogonal transform in Radon are generally designed in the integer-order space. As to be seen later, this restriction will lead to a limited flexibility and discriminability of the representation, especially the spatial information is neglected.

\subsection{Fractional-order Moments}

The research of Fractional-order Moments (FM) arises mainly from the information suppression problem. Mathematically, the distribution of zeros of the moment basis functions is related to the description emphasis over image plane \cite{ref33}. Such suppression means that moments unnecessarily emphasize non-discriminative parts of image, resulting from the biased distributions of zeros. 

In pattern recognition, the information suppression problem was first demonstrated empirically by Kan et al. \cite{ref34}, through comparing the biased ZM with the unbiased OFMM. As a pioneering work, Hoang et al. \cite{ref35} proposed the first class of FM called Generic Polar Harmonic Transforms (GPHT). It can control the zero distributions by changing fractional-order parameter, thereby being useful for solving suppression problem and extracting local feature. Following this paper, a list of FM was formalized in Cartesian or polar coordinates, e.g., fractional-order Legendre moments \cite{ref36}, fractional-order OFMM \cite{ref37}, fractional-order ZM \cite{ref38}, and fractional-order Chebyshev moments \cite{ref39}. More recently, Yang et al. \cite{ref23} provided a generic Fractional-order JFM (FJFM) and a systematic discussion on the time-frequency discriminability. They found that an alternative representation path is to combine low-order basis functions with complementary distributions of zeros, thus capturing complementary image information, and alleviating the contradiction between the robustness and discriminability. Among them, the GPHT and FJFM are more generic definitions serving as unified mathematical tools for the research of FM. 

Note that all above FM are designed in the original image domain; while in the Radon domain, their definition, noise robustness, rotation invariance, time-frequency discriminability, and application have not yet been explored.

\subsection{Image Denoising}

Unlike such efforts for improving the robustness/invariance of the representation itself, the restoration strategy, e.g. image denoising, is another long-lasting research path \cite{ref63}.

For estimating a maximally approximated version of the original scene from noisy observation, prior knowledge about noise and scene is desired due to the ill-posed nature of this problem \cite{ref10, ref11}. In the hand-crafted era, the design of denoising algorithms was mainly motivated by the distribution patterns of noise and scene signals. A milestone is DM3D \cite{ref64}, with explicitly representations for local smoothness and nonlocal self-similarity of natural image in the spatial domain, as well as statistical properties of noise in the frequency domain. In the deep-learning era, more researchers believe that such patterns can be implicitly modeled by deep representations, covering a full range of complex and realistic distribution priors. In this regard, deep networks such as DnCNN \cite{ref52}, FFDNet \cite{ref65}, and SRMD \cite{ref66} have been designed to adaptively learn the mapping from noisy images to clean images, achieving the current state-of-the-art results from the human perspective.

However, for both hand-crafted and deep denoising, they are typically evaluated by the similarity between the resulting image and the ground-truth image \cite{ref63}, i.e., quality assessment or user scoring. In the community, denoising preprocessing and subsequent automatic task are rarely optimized and tested in a combination, leading to a certain research gap \cite{ref12, ref13}. Specifically, such gap has the following forms: 1) applicability issues due to the lack of noise prior; 2) accuracy and robustness issues due to the errors introduced by the unstable process; and 3) efficiency issues due to the preprocessing.

\subsection{Robust Visual Tasks}

The applications of deep learning techniques are expanding into trustworthy scenarios, e.g., deepfake detection \cite{ref67}, medical diagnostics \cite{ref68}, self-driving cars \cite{ref69}, and biometrics \cite{ref70}. Here, the robustness is crucial \cite{ref71} for trustworthy, meaning that the performance of system is stable for intra-class variations on the input.

In this paper, a class of robust visual tasks that we are mainly interested in is visual-content security and forensics \cite{ref72}. It is an emerging topic that aims to guarantee the  authenticity and controllability of visual data under adversary assumptions. In general, the solutions are divided into two paths: active and passive. Active ones rely on speciﬁc information embedding \cite{ref73} or extracting \cite{ref74} for the image prior to distribution. With the side information, such methods are capable of verifying image with almost any type of manipulations. Yet, they require additional implementation costs and are not applicable to distributed images. In contrast, passive solutions do not rely on such prior processing. They work exclusively on the given image itself, by finding the clues at the digital \cite{ref75}, physical \cite{ref76}, or semantic \cite{ref77} level, where such clues are inherent in the natural images or introduced inevitably by certain manipulations. As for the downside, passive methods are unsatisfactory in robustness and generalization, due to the lack of side information.

\section{Foundations}

For the sake of completeness, we briefly remind some foundations of Radon transform.

\begin{definition} \textbf{(Radon transform).}
	 The Radon transform of an image function $f$, denoted as ${{\cal R}_f}$, is defined as a line integral along a straight line $L$ \cite{ref40}:
		\begin{equation}
			\begin{split}
			&{{\cal R}_f}(r,\theta ) = \int_{L(r,\theta )} {f(x,y)dxdy}\\ 
			&= \int\limits_{ - \infty }^\infty  {\int\limits_{ - \infty }^\infty  {f(x,y)\delta (r - x\cos \theta  - y\sin \theta )dxdy} },
			\end{split}
		\end{equation}
	where the line $L(r,\theta ) = \{ (x,y) \in \mathbb{R}^2 \mathrm{\;s.t.\;} r - x\cos \theta  - y\sin \theta  = 0\} $ is with distance parameter $r$ (w.r.t. the origin) and angle parameter $\theta $ (w.r.t. the $y$-axis); $\delta $ is the Kronecker delta function: $\delta (\cdot) = [\cdot = 0]$.
\end{definition}

In Fig. 1, we give a high-level intuition for the performance of the Radon transform, with the geometric illustration of above involved parameters/terms.

\begin{proposition} \textbf{(Rotation covariance of Radon transform).}
	Suppose ${f_R}$ is a rotated version of image ${f}$ by an angle $\varphi $ about the center, the Radon transform of ${f_R}$ is a circular shift of the ${{\cal R}_f}$ in parameter $\theta $ with the same angle $\varphi $ \cite{ref41}:
	\begin{equation}
		{{\cal R}_{{f_R}}}(r,\theta ) = {{\cal R}_f}(r,\theta  + \varphi ),
	\end{equation}
    implying the covariance w.r.t. rotation.
\end{proposition}

\begin{proposition} \textbf{(Robustness to additive white noise of Radon transform).}
	Suppose ${f_\eta }$ is a noisy version of image $f$ corrupted by additive white noise $\eta $: ${f_\eta } = f + \eta $, the Radon transforms of ${f_\eta }$ and $f$ are identical under \emph{continuous-domain} assumption as \cite{ref42}:
	\begin{equation}
		{{\cal R}_{{f_\eta }}}(r,\theta ) = {{\cal R}_f}(r,\theta ) + {{\cal R}_\eta }(r,\theta ) = {{\cal R}_f}(r,\theta ),
	\end{equation}
	where ${{\cal R}_\eta }(r,\theta ) = 0$ as the Radon transform of additive noise is proportional to the mean value of the noise. Note that, as for  \emph{discrete-domain} versions in practice, such identity will not hold. Suppose ${f_\eta }$ and $f$ are sampled and quantized digital images: $f$ is with mean $\mu $, $\eta $ is with variance ${\sigma ^2}$, and both is in size of $N \times M$ (pixels), the SNRs of noisy image ${f_\eta }$ and its Radon projection along a $\theta $, i.e., ${{\cal R}_{{f_\eta }}}(\cdot,\theta )$, exhibit the following relation \cite{ref43}:
	\begin{equation}
		\mathrm{SNR}_{{{\cal R}_{{f_\eta }}}(\cdot,\theta )} = \mathrm{SNR}_{{f_\eta }} + \frac{{{\mu ^2}(c(\theta ) - 1)}}{{{\sigma ^2}}},
	\end{equation}
	where $c(\theta )$ is a term w.r.t. Radon projection, roughly in the interval $[\min (N,M),\max (N,M)]$.
\end{proposition}

This proposition indicates that the increment in SNR after Radon projection is ${\mu ^2}(c(\theta ) - 1)/{\sigma ^2}$, which is typically large in practice, meaning a strong noise robustness. Here, a pertinent example is given for an intuitive perception.

\begin{example}
	Assuming that the pixel mean $\mu  = 0.5$, noise variance ${\sigma ^2} = 0.1$ (both under grayscale normalization), and image size $N = M = 256$, the increment in SNR is ${\mu ^2}(c(\theta ) - 1)/{\sigma ^2} \approx 637.5$.
\end{example}

\begin{figure}[!t]
	\centering
	\subfigure[]{\includegraphics[scale=0.4]{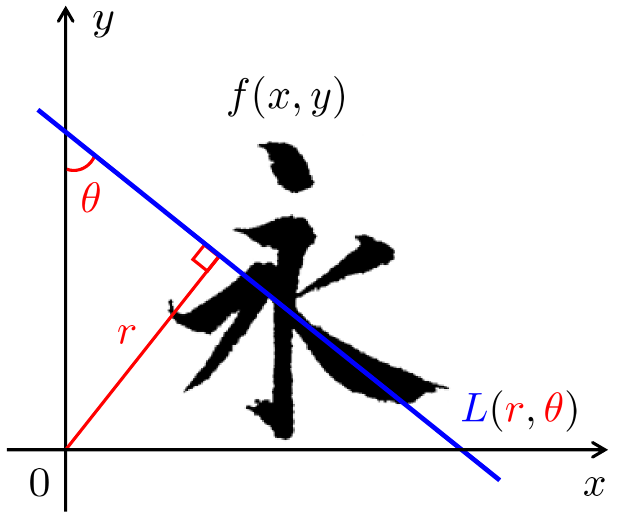}}
	\subfigure[]{\includegraphics[scale=0.4]{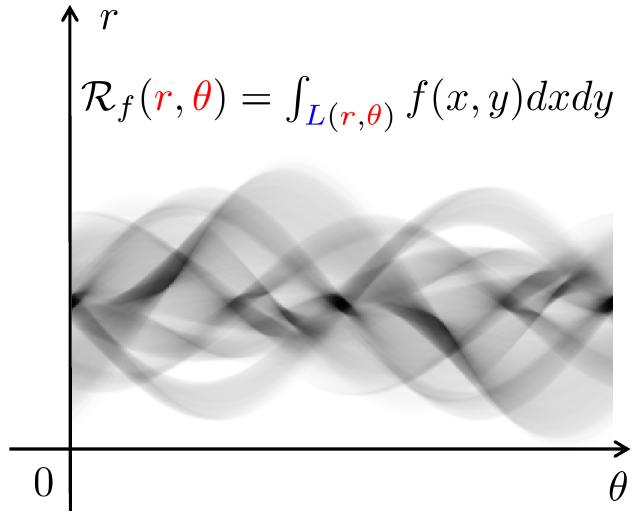}}
	\centering
	\caption{Illustration for the Radon transform. (a) The image function $f(x,y)$ and a straight line $L(r,\theta )$ with distance $r$ to the origin and angle $\theta $ to the $y$-axis. (b) The line integral along such straight line, $\int_{L(r,\theta )} {f(x,y)dxdy} $, as the corresponding Radon transform ${{\cal R}_f}(r,\theta )$.}
\end{figure}

\section{Fractional-order Moments in Radon:\\ An Implicit Path}

This section provides the implicit path for constructing our Fractional-order Moments in Radon (FMR). We start with the mathematical definition of FMR and then delve into the analysis of its derived beneficial properties. Finally, we design an efficient implementation strategy for FMR in a realistic computing environment.

\subsection{Implicit Definition}

Mathematically, the general theory of this work is based on the following definition, i.e., an inner product for the basis function and the Radon transform of the image function.

\begin{definition} \textbf{(Fractional-order moments in Radon).}
	For an image function $f$, the fractional-order moments in Radon of $f$ is defined as the following inner product:
	\begin{equation}
		\left< {{\cal R}_f},V_{nm}^\alpha \right>  = \iint_{D}{{{[V_{nm}^\alpha (r,\theta )]}^ * }{{\cal R}_f}(r,\theta )rdrd\theta },
	\end{equation}
	where ${{\cal R}_f}$ is the Radon transform of $f$ defined in (1), $V_{nm}^\alpha $ is the basis function with order parameter $(n,m) \in {\mathbb{Z}^2}$ and fractional parameter $\alpha \in {\mathbb{R}^+}$ on the domain $D \in {\mathbb{R}^2}$, and $ * $ denotes the complex conjugate. In this paper, two beneficial constraints, i.e., \emph{orthogonality} and \emph{in-form rotation invariance}, are imposed on the definition of basis functions, leading to an explicit form:
	\begin{equation}
		V_{nm}^\alpha (r,\theta ) = R_n^\alpha (r){A_m}(\theta ),
	\end{equation}
	with \emph{angular basis function} ${A_m}(\theta ) = \exp (\bm{j}m\theta )$ ($\bm{j} = \sqrt { - 1} $) and \emph{radial basis function} $R_n^\alpha (r)$ satisfying the weighted orthogonality condition: $\int\limits_0^1 {R_n^\alpha (r){{[R_{n'}^\alpha (r)]}^*}rdr}  = \frac{1}{{2\pi }}{\delta _{nn'}}$ over a unit-disk domain $D = \{ (r,\theta ):r \in [0,1],\theta  \in [0,2\pi )\} $.
\end{definition}

For the above definition, the explicit form of $R_n^\alpha$ is not yet given. Ideally, it can be formed from almost any kinds of orthogonal functions in mathematical literature (see \cite{ref15} for a survey). In order to benefit the representation and implementation while not losing the generality on the following discussion, we introduce two classes of fractional-order orthogonal functions as radial basis functions, based on harmonic and polynomial functions respectively.

\begin{definition} \textbf{(Harmonic radial basis function).}
	The \emph{complex exponential functions} \cite{ref35} in Fourier analysis can be employ for defining $R_n^\alpha$ as:
	\begin{equation}
		R_n^\alpha (r) = \sqrt {\frac{{\alpha {r^{\alpha  - 2}}}}{{2\pi }}} \exp (\bm{j}2n\pi {r^\alpha }),
	\end{equation}
	which directly satisfies the weighted orthogonality condition in Definition 2 for all possible $\alpha $.
\end{definition}

\begin{definition} \textbf{(Polynomial radial basis function).}
	A class of \emph{classical orthogonal polynomials}, Jacobi polynomials \cite{ref23}, can be employ for defining $R_n^\alpha$  as:
	\begin{equation}
		\begin{split}
		R_n^\alpha (p,q,r) = \sqrt {\frac{{\alpha {r^{\alpha q - 2}}{{(1 - {r^\alpha })}^{p - q}}(p + 2n) \Gamma (q + n) n!}}{{2\pi \Gamma (p + n) \Gamma (p - q + n + 1)}}} \\
		\times \sum\limits_{k = 0}^n {\frac{{{{( - 1)}^k}\Gamma (p + n + k){r^{\alpha k}}}}{{k!(n - k)!\Gamma (q + k)}}},
		\end{split}
	\end{equation}
	where $n \in \mathbb{N}$ and the polynomial parameters $p,q \in \mathbb{R}$ must fulfill: $p - q >  - 1$, $q > 0$; similarly, the above definition directly satisfies the weighted orthogonality condition in Definition 2 for all possible $\alpha $, $p$, and $q$.
\end{definition}

\emph{Recalling the implicit definition.} The FMR with harmonic/polynomial radial basis function is termed as harmonic/polynomial FMR throughout this paper. With the Definitions 1 $\sim$ 4, we formulate the most basic components in this paper. More specifically, a generic representation framework in the Radon domain under the constraints of orthogonality and in-form rotation invariance has been designed, with also optional harmonic or polynomial radial basis functions.

\begin{figure}[!t]
	\centering
	\includegraphics[scale=0.4]{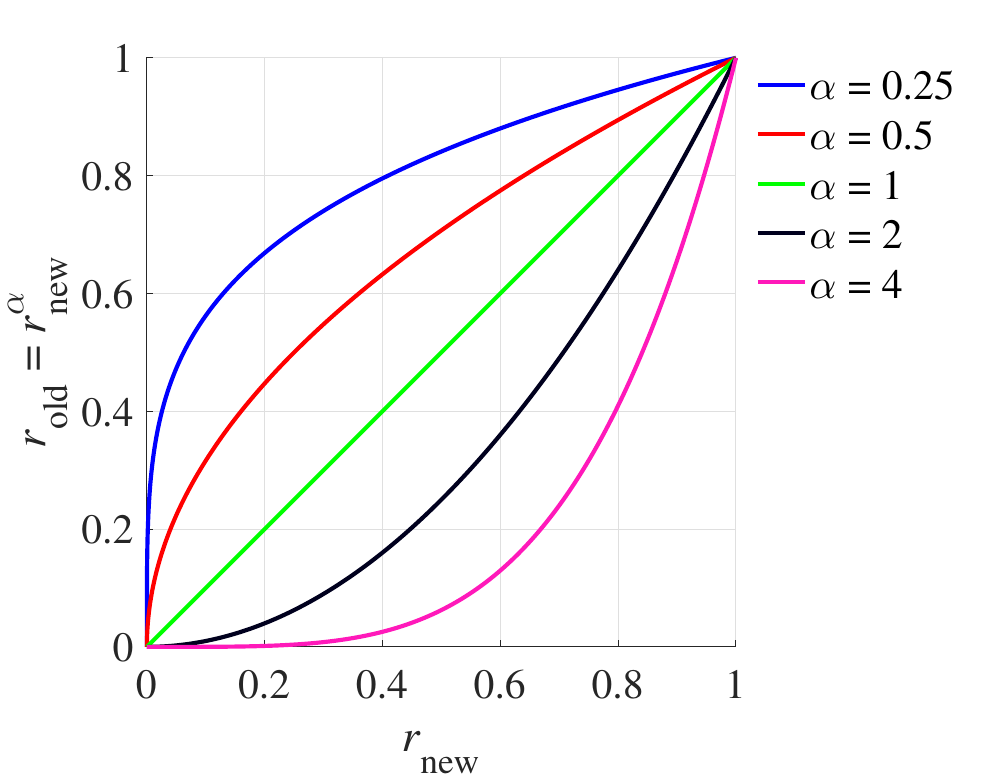}
	\centering
	\caption{Illustration for the effect of fractional parameter $\alpha  \in {\mathbb{R}^ + }$ on a pair of radial independent variables: ${r_{{\rm{new}}}} \in [0,1]$ and ${r_{{\rm{old}}}} = {r_{{\rm{new}}}}^\alpha  \in [0,1]$.}
\end{figure}

\subsection{Property}

Let us discuss some important properties derived from above definitions, covering generic nature, rotation invariance, noise robustness, and time-frequency discriminability.

\begin{property} \textbf{(Generic nature).}
	By taking $\alpha $ in the above definitions as a parameter, the FMR is a mathematical generalization of the integer-order orthogonal moments in Radon with harmonic or polynomial radial basis functions. A specific Radon-domain representation can be obtained by setting a specific value of $\alpha $.
\end{property}

\emph{Remark.} In general, the existing related works, e.g., \cite{ref24}, are special cases for our FMR with fixed parameter $\alpha  = 1$ or $\alpha  = 2$. It is also remarkable that, for a class of FMR representations obtained by changing the value of $\alpha $, members of this class share common robustness/invariance benefits (i.e., Properties 2 and 3) while also being complementary in discriminative information (i.e., Property 4).

\begin{property} \textbf{(Rotation invariance).}
	Suppose ${f_R}$ is a rotated version of image $f$ by an angle $\varphi $ about the center, there must be a function ${\cal I}$ other than constant functions such that:
	\begin{equation}
		{\cal I}(\{ \left< {{\cal R}_{{f_R}}},V_{nm}^\alpha \right> \} ) \equiv {\cal I}(\{ \left< {{\cal R}_f},V_{nm}^\alpha \right> \} ),
	\end{equation}
	i.e., satisfying the rotation invariance, for all possible input images and FMR parameters.
\end{property}

\emph{Proof.} With the Proposition 1 and Definition 2, it can be checked that the FMR of rotated image $\left< {{\cal R}_{{f_R}}},V_{nm}^\alpha \right>$ and the FMR of original image $\left< {{\cal R}_f},V_{nm}^\alpha \right>$ differ by only a phase term related to the rotation angle $\varphi $, as follows:
\begin{equation}
	\begin{split}
		&\left< {{\cal R}_{{f_R}}}(r,\theta ),V_{nm}^\alpha (r,\theta ) \right>\\
		&= \left< {{\cal R}_f}(r,\theta  + \varphi ),V_{nm}^\alpha (r,\theta ) \right> \\
		&=\left< {{\cal R}_f}(r,\theta '),V_{nm}^\alpha (r,\theta ' - \varphi ) \right> \\ 
		&=\left< {{\cal R}_f}(r,\theta '),V_{nm}^\alpha (r,\theta ') \right> {[{A_m}( - \varphi )]^*}\\
		&=\left< {{\cal R}_f},V_{nm}^\alpha  \right> {A_m}(\varphi ),
	\end{split}
\end{equation}
where $\theta ' = \theta  + \varphi $. Note that the first and second passes are true as ${{\cal R}_{{f_R}}}(r,\theta ) = {{\cal R}_f}(r,\theta  + \varphi )$ and ${A_m}(\theta ) = \exp (\bm{j}m\theta )$, respectively. From the above insight, we should look for such function — properly eliminating the effect of the extra phase term ${A_m}(\varphi )$ — as the ${\cal I}$. For example, a naive approach is to directly remove the phase information of the representation with magnitude operator:
\begin{equation}
	{\cal I}(\left< {{\cal R}_ \cdot },V_{nm}^\alpha \right> ) \triangleq | \left< {{\cal R}_ \cdot },V_{nm}^\alpha \right> |,
\end{equation}
obviously satisfying $| \left< {{\cal R}_{{f_R}}},V_{nm}^\alpha \right> | = | \left< {{\cal R}_f},V_{nm}^\alpha \right> {A_m}(\varphi )| = | \left< {{\cal R}_f},V_{nm}^\alpha \right>|$. For an information-preserving representation, another approach generalizes such magnitude operator to a flexible phase cancelation as \cite{ref44}:
\begin{equation}
	{\cal I}( \left< {{\cal R}_ \cdot },V_{nm}^\alpha \right> ) \triangleq  \prod\limits_{i = 1}^L {{{( \left< {{\cal R}_ \cdot },V_{n{m_i}}^\alpha \right> )}^{{k_i}}}},
\end{equation}
where $L \ge 1$, ${m_i},{k_i} \in  \mathbb{Z}$, $i = 1,...,L$, such that $\sum\limits_{i = 1}^L {{m_i}{k_i} = 0} $.

\qed

For more details on such phase cancelation, we address the reader to \cite{ref44}.

\begin{property} \textbf{(Robustness to additive white noise).}
	Suppose ${f_\eta }$ is a noisy version of image $f$ corrupted by additive white noise $\eta $: ${f_\eta } = f + \eta $, the FMR is more robust than the counterpart in original image domain under the sense of $l$-norm distance of representation variations:
	\begin{equation}
		\begin{split}
		&||\{\left< {{\cal R}_{{f_\eta }}},V_{nm}^\alpha \right> \}  - \{ \left< {{\cal R}_f},V_{nm}^\alpha\right> \} ||_l \\
		&\le ||\{\left< {f_\eta },V_{nm}^\alpha\right> \}  - \{\left< f,V_{nm}^\alpha \right> \} ||_l,
		\end{split}
	\end{equation}
	and the equality holds if and only if $\eta  = 0$.
\end{property}

\emph{Proof.} In the continuous domain, we can rewrite the left side of (13) by the Proposition 2 and Definition 2 as:
\begin{equation}
	\begin{split}
	&||\{  \left< {{\cal R}_{{f_\eta }}},V_{nm}^\alpha  \right> \}  - \{  \left< {{\cal R}_f},V_{nm}^\alpha  \right> \} ||_l\\
	&=||\{  \left< {{\cal R}_{{f_\eta }}}-{{\cal R}_f},V_{nm}^\alpha  \right> \} ||_l\\
	&= ||\{  \left<0,V_{nm}^\alpha  \right> \} ||_l = 0,
	\end{split}
\end{equation}
where the first pass is true as the linearity of inner product and the second pass is true as (3); similarly, the right side of (13) can be rewritten as:
\begin{equation}
	\begin{split}
		&||\{  \left< {f_\eta },V_{nm}^\alpha  \right> \}  - \{  \left< f,V_{nm}^\alpha  \right> \} ||_l\\
		&= ||\{  \left< {f_\eta - f },V_{nm}^\alpha  \right> \} ||_l\\
		&= ||\{  \left< \eta ,V_{nm}^\alpha  \right> \} ||_l \ge 0,
	\end{split}
\end{equation}
where we assert that $||\{  \left< \eta ,V_{nm}^\alpha  \right> \} ||_l \ge 0 $  and the equality holds if and only if $\eta=0$. Without loss of generality, we assume that there exists $\eta \neq 0$, such that $||\{ \left<\eta ,V_{nm}^\alpha \right>  \} ||_l = 0 $. Based on this assumption, it is known that $\eta \notin \{ V_{nm}^\alpha \}$, due to $||\{ \left<\eta,V_{nm}^\alpha\right> \} ||_l = ||\delta ||_l \ne 0$ if $\eta \in \{ V_{nm}^\alpha \}$. Taken together, the above assumption means that $\eta$ is a nonzero element that does not belong to the set of basis functions, while being orthogonal to any element in this set for guaranteeing $||\{ \left< \eta ,V_{nm}^\alpha \right>  \} ||_l = 0 $. This obviously conflicts with the completeness and orthogonality of the basis functions in a Hilbert space. Therefore, this assumption does not hold. It can be noted that the correctness of (14) and (15) implies in fact the correctness of the original property (13).

\qed

Although the ideal identity (14) will no longer hold for the discrete domain, the robustness of FMR is still guaranteed because the SNR of noisy image ${f_\eta }$ is significantly lower than and the SNR of its Radon transform ${{\cal R}_{{f_\eta }}}$, as illustrated in (4).

\begin{property} \textbf{(Time-frequency discriminability).}
	The frequency and spatial properties of the represented image information by FMR can be controlled with the order parameter $(n,m)$ and fractional parameter $\alpha $, respectively.
\end{property}

\emph{Remark.} In the community of moments and moment invariants, researchers have found that such frequency and spatial properties depend on the number and location of zeros of the basis functions, respectively. As for our FMR, the number and location of zeros can be adjusted explicitly with $(n,m)$ and $\alpha $ respectively, which is the core of achieving time-frequency discriminability.

\begin{itemize}
	\item Regarding the number, it is proportional to the $n$ and $m$ in the radial $R_n^\alpha$ and angular ${A_m}$, respectively, due to the intrinsic nature of harmonic and polynomial functions.
	\item Regarding the location, in the radial $R_n^\alpha$, it is biased towards $0$ when $\alpha  < 1$, generally uniform when $\alpha  = 1$, and biased towards $1$ when $\alpha  > 1$, where the more deviation of $\alpha $ from $1$ is, the more biased the distribution of zeros is \cite{ref23}. The factor underlying this property is that, when $\alpha $ is less/greater than $1$, ${r^\alpha }$ is constantly greater/less than ${r^1}$ over the domain $(0,1)$, which in turn corresponds to a left/right shift of zeros w.r.t. $\alpha  = 1$, as illustrated in Fig. 2.
\end{itemize}

By the way, we provide a comment on the notation $V_{nm}^\alpha$: the superscript and subscript denote the parameters for spatial and frequency domains, respectively.

For the practical intuitions of such time-frequency discriminability, we illustrate the phases of $V_{nm}^\alpha$ in Fig. 3, under different parameter settings: $n = 5$, $m = \{ 2,4,8\} $, and $\alpha  = \{ 1/2,1,2\} $. As expected from above theory, changing $m$ will change the number of zeros in the angular direction of the basis functions, which in turn corresponds to a change in the frequency properties. Here, a similar phenomenon also holds for $n$ but is not shown in Fig. 3. As for the newly introduced $\alpha $, the change of its value will change the distribution of zeros in the radial direction of the basis functions, which in turn corresponds to a change in the spatial properties. Such time-frequency discriminability is not available in the previous integer-order method, which only corresponds to the case of $\alpha  = 1$.

\begin{figure}[!t]
	\centering
	\includegraphics[scale=0.4]{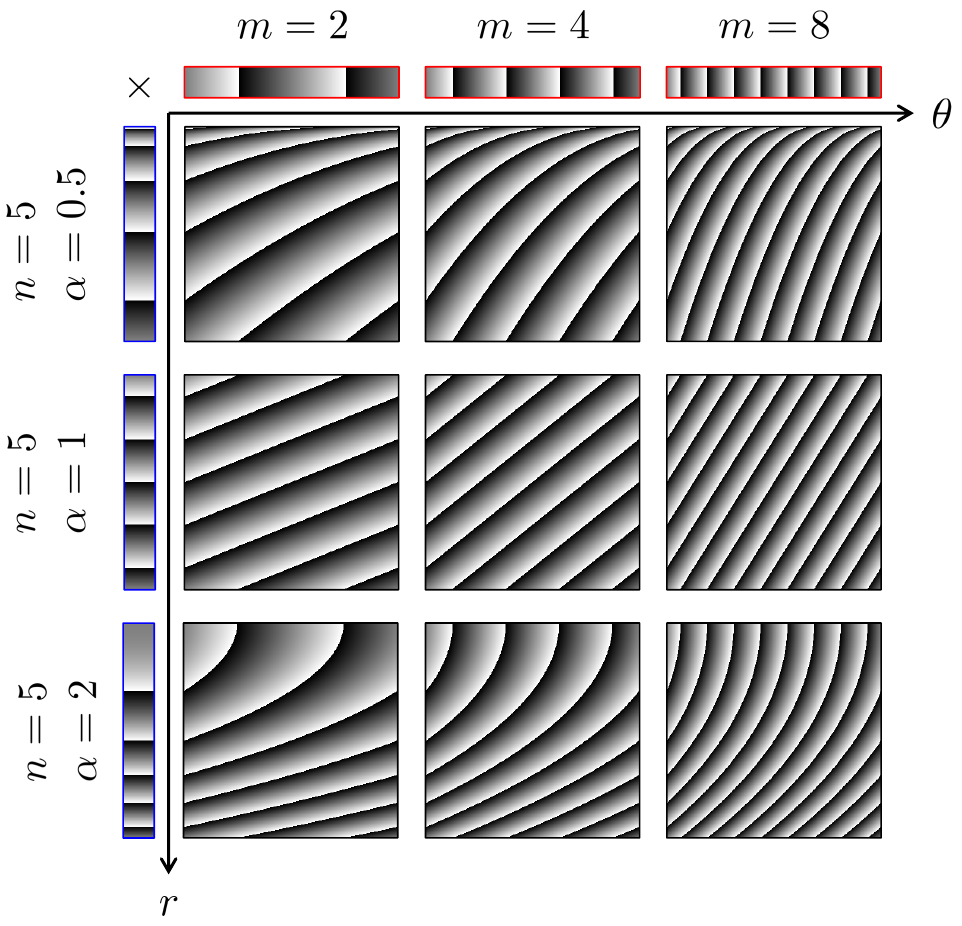}
	\centering
	\caption{Illustration for the time-frequency discriminability. Under different parameter settings, the phases for the angular basis function ${A_m}(\theta )$, the harmonic radial basis function $R_n^\alpha (r)$ in (7), and the 2D basis function $V_{nm}^\alpha (r,\theta )$ (as the product of the two) are plotted with red, blue, and black borders, respectively. Note that $n$ and $m$ encode the radial and angular frequency properties, respectively, while the newly introduced $\alpha $ encodes the radial spatial properties which is not available in the previous integer-order approach (i.e., with $\alpha  = 1$).}
\end{figure}

\emph{Recalling the property.} It is worth highlighting the distinctiveness of our FMR w.r.t. some theoretically relevant methods. As illustrated in Table 1, our FMR is a more comprehensive framework with all above Properties 1 $\sim$ 4, compared to the image representations in the four typical frameworks. Here, the moments in image do not have Property 3, the integer-order moments/transformations in Radon do not have Properties 1 and 4, while some other Radon representations further do not have Property 2.

\begin{figure*}[!t]
	\centering
	\includegraphics[scale=0.4]{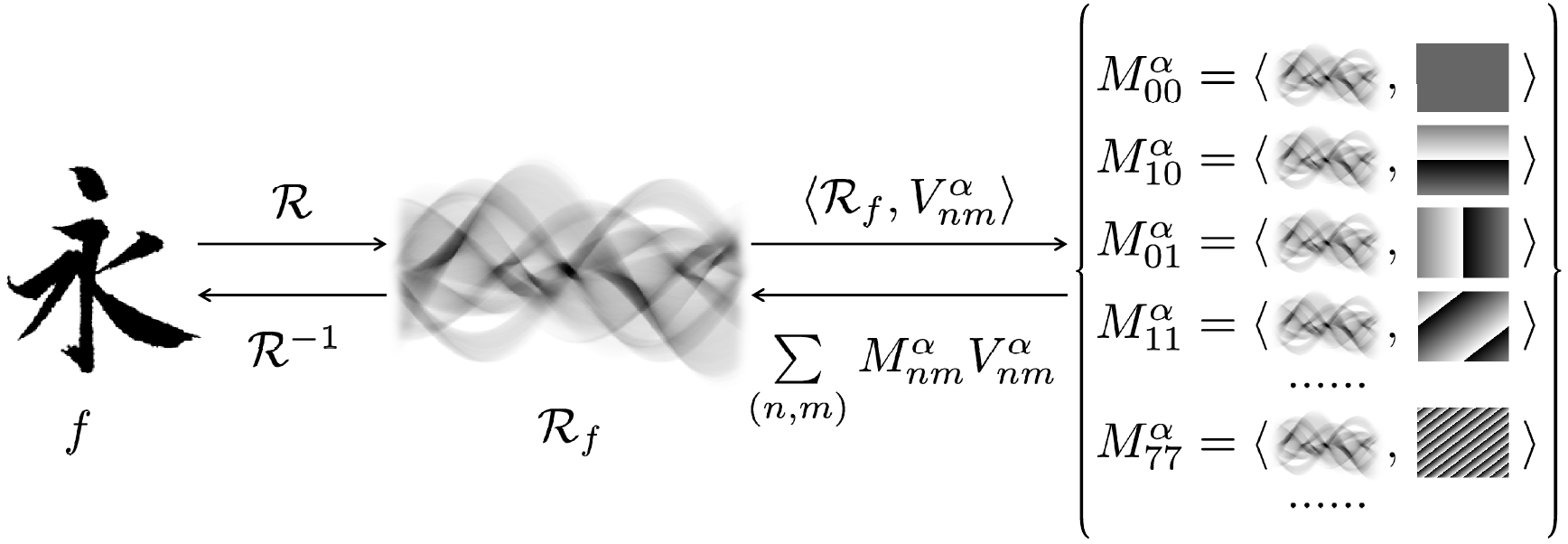}
	\centering
	\caption{Illustration for the implementation of FMR. The original image $f$ is first projected into the Radon space as ${{\cal R}_f}$. Then, the inner product of ${{\cal R}_f}$ and the basis function $V_{nm}^\alpha $ with different parameters $(\alpha ,n,m)$ is computed as FMR $M_{nm}^\alpha  =  \left< {{\cal R}_f},V_{nm}^\alpha \right> $. Here, the computation of the inner product and the estimation of the basis function can be efficiently executed by the Fourier and recursive based strategies, respectively. Note that the above transformation is invertible.}
\end{figure*}

\subsection{Implementation}

The definition and the derived properties of FMR have been described in Sections 4.1 and 4.2, both under a continuous-domain assumption. Now, we will focus on the specific implementations of FMR on digital images, considering their accuracy and efficiency.

For the FMR with harmonic radial basis function of Definition 3 (i.e., harmonic FMR), we introduce the following Fourier-based definition, which can further lead to a computationally efficient and numerically stable solution.

\begin{theorem} \textbf{(Harmonic FMR by Fourier transform: Analytical formula).}
	The harmonic FMR with Definitions 2 and 3 can be rewritten into a Fourier transform form \cite{ref45}, as follows:
	\begin{equation}
		\left< {{\cal R}_f},V_{nm}^\alpha \right>  = 2\pi {\cal F}({\cal S}(\gamma ,\vartheta )),
	\end{equation}
	where $\gamma  = {r^\alpha }$, $\vartheta  = \theta /2\pi $, and ${\cal S}(\gamma ,\vartheta ) = \sqrt{{\gamma ^{{\alpha  \mathord{\left/ {\vphantom {\alpha  2}} \right. \kern-\nulldelimiterspace} 2} - 1}}/2\pi \alpha }\times{{\cal R}_f}(\underbrace {{\gamma ^{\frac{1}{\alpha }}}}_r,\underbrace {2\pi \vartheta }_\theta )$.
\end{theorem}

\emph{Proof.} The proof is given in Appendix A, available in the online supplemental material. \qed

We give the following lemma as a numerically efficient way to conduct above analytical formula (16).

\begin{lemma} \textbf{(Harmonic FMR by Fourier transform: Numerical formula).}
	With sampling and zero-order approximation for both Fourier basis and ${\cal S}(\gamma ,\vartheta )$, the analytical formula (16) can be rewritten into an approximate discrete form \cite{ref45}, as follows:
	\begin{equation}
		\left< {{\cal R}_f},V_{nm}^\alpha \right>  \simeq \frac{{2\pi }}{{{M^2}}}{\cal F}[{\cal S}[u,v]],
	\end{equation}
	where ${\cal S}[u,v]\mathop  \triangleq {\cal S}({\gamma _u},{\vartheta _v}) = \sqrt {\frac{{{\gamma _u}^{{\alpha  \mathord{\left/ {\vphantom {\alpha  2}} \right. \kern-\nulldelimiterspace} 2} - 1}}}{{2\pi \alpha }}} {{\cal R}_f}({r_u},{\theta _v})$ (the square brackets suggest the use of discrete variables) under sampling:
	\begin{equation}
	\left\{ {\begin{array}{l}
				{{\gamma _u} = r_u^\alpha  = \frac{u}{M}}\\
				{{\vartheta _v} = \frac{{{\theta _v}}}{{2\pi }} = \frac{v}{M}}
		\end{array}} \right.,
	\left\{ {\begin{array}{l}
			{\Delta {\gamma _u} = \frac{1}{M}}\\
			{\Delta {\vartheta _v} = \frac{1}{M}}
	\end{array}} \right., 
	\left\{ {\begin{array}{l}
		{{r_u} = {{(\frac{u}{M})}^{\frac{1}{\alpha }}}}\\
		{{\theta _v} = \frac{{2\pi v}}{M}}
		\end{array}} \right.
	\end{equation}
	where $(u,v) \in {\{ 0,1,...,M - 1\} ^2}$ and $M$ is a constant related to the sampling rate. In our practice, the computationally efficient Fast Fourier Transform (FFT) is used for the implementation of (18).
\end{lemma}

\emph{Remark.} With the analytical Theorem 1 and corresponding numerical Lemma 1, our harmonic FMR is able to be calculated in a more stable and fast manner, than the direct implementation from Definitions 2 and 3.

\begin{itemize}
	\item \emph{Stability.} The direct implementation and the above Fourier-based implementation are actually built in Cartesian and polar coordinate systems, respectively. Due to a well-known relationship $dxdy = rdrd\theta$, the radial parts for such two implementations have different boundedness properties: ${\lim _{(x,y) \to (0,0)}}|R_n^\alpha (\sqrt {{x^2} + {y^2}} )| = \infty $ for  $\alpha  \in (0,2)$ and ${\lim _{r \to 0}}|r \cdot R_n^\alpha (r)| = 0$ for $\alpha  \in (0, + \infty )$, respectively. Here, the proof is straightforward based on the Definition 3. Clearly, the Fourier-based implementation has better stability at $r \simeq 0$.
	\item \emph{Efficiency.} Considering the calculation for a set of $\left< {{\cal R}_f},V_{nm}^\alpha \right>$ with $(n,m) \in {\{  - K,...,0,1,...,K\} ^2}$ and a fixed $\alpha $, where the digital image of size $N \times N$ and the sampling parameter $M \propto N$. The direct implementation requires ${\cal{O}}({N^2}{(2K + 1)^2}) = {\cal{O}}({N^2}{K^2})$ multiplications as the multiplying point by point over the Cartesian sampling. The Fourier-based implementation requires ${\cal{O}}({M^2}\log M) = {\cal{O}}({N^2}\log N)$ multiplications by the FFT, where $K$ is surprisingly no role in the complexity, meaning a constant-time calculation w.r.t. the size of order set. Owing to the significantly different growth properties of ${K^2}$ and $\log N$, the Fourier-based implementation has better efficiency.
\end{itemize}

For the FMR with polynomial radial basis function of Definition 4 (i.e., polynomial FMR), we introduce the following recursive definition, which can further lead to a computationally efficient and numerically stable solution.

\begin{theorem} \textbf{(Polynomial FMR by recursive relations: Analytical formula).}
	The polynomial radial basis function with Definition 4 can be rewritten into a recursive form \cite{ref23}, as follows:
	\begin{equation}
		\begin{split}
		R_n^\alpha (p,q,r) = \sqrt {\frac{{(p + 2n)\alpha {r^{\alpha q - 1}}{{(1 - {r^\alpha })}^{p - q}}}}{r}}\\
		\times{C_n}(p,q)P_n^\alpha (p,q,r),
		\end{split}
	\end{equation}
	where the term $P_n^\alpha (p,q,r)$ is defined recursively as:
	\begin{equation}
		P_n^\alpha  = ({L_1}{r^\alpha } + {L_2})P_{n - 1}^\alpha  + {L_3}P_{n - 2}^\alpha, n \ge 2, 
	\end{equation}
	with factors: ${L_1} =  - (2n + p - 1)(2n + p - 2)/(n(q + n - 1))$, ${L_2} = (p + 2n - 2) + {L_1}(n - 1)(q + n - 2)/(p + 2n - 3)$, and ${L_3} = (p + 2n - 4)(p + 2n - 3)/2 + {L_1}(q + n - 3)(n - 2)/2 - (p + 2n - 4){L_2}$ also with initial values: ${P_0} = \Gamma (p)/\Gamma (q)$ and ${P_1} = (1 - {r^\alpha }(p + 1)/q)  \Gamma (p + 1)/\Gamma (q)$; the term ${C_n}(p,q)$ is defined recursively as:
	\begin{equation}
		{C_n} = \sqrt {\frac{{n(q + n - 1)}}{{(p + n - 1)(p - q + n)}}} {C_{n - 1}}, n \ge 1, 
	\end{equation}
	with initial value: ${C_0} = \sqrt {\Gamma (q)/(\Gamma (p)\Gamma (p - q + 1))} $.
\end{theorem}

Note that the proof of Theorem 2 is trivial but relies solely on the properties $a! = a \cdot (a - 1)!$ and $\Gamma (a) = a \cdot \Gamma (a - 1)$. In addition, it is straightforward to extend the above analytical formulas to numerical ones, i.e., such recursive relations holds directly on the sampling. Therefore, the proof and numerical version of Theorem 2 will not be covered in this paper.

\emph{Remark.} With the Theorem 2, our polynomial FMR is able to be calculated in a more stable and fast manner, than the direct implementation from Definitions 2 and 4.

\begin{itemize}
	\item \emph{Stability.} The factorial/gamma terms in (8) are the main factor leading to the numerical instability, e.g., the stability of $a!$ collapses when $a > 21$ for a modern floating-point arithmetic. By comparing the direct implementation from (8) and above recursive implementation, we can note that factorial/gamma terms for large numbers are avoided in the recursive one, implying a better stability.
	\item \emph{Efficiency.} Considering the calculation for a $R_n^\alpha (p,q,r)$ with fixed $n$ and $\alpha $ over a digital image of size $N \times N$. The direct implementation requires ${\cal{O}}(n{N^2})$ additions as the $n$-time summation over $N \times N$ points in (8). The recursive implementation from the calculated $R_{n - 1}^\alpha $ and $R_{n - 2}^\alpha $ requires only ${\cal{O}}({N^2})$ additions for the one summation. As for a set of $R_n^\alpha (p,q,r)$ with $n \in {\{ 0,1,...,K\} ^2}$ and a fixed $\alpha $, the addition complexities of direct and recursive implementations are ${\cal{O}}({K^2}{N^2})$ and ${\cal{O}}(K{N^2})$, respectively. Clearly, the recursive implementation has better efficiency.
\end{itemize}

\emph{Recalling the implementation.} It is worth give a high-level intuition for the entire implementation. As illustrated in Fig. 4, the original image is first projected by the Radon transform (Section 3) and subsequently decomposed into orthogonal moments (Section 4). For the computation of the inner product and the estimation of the basis function in the above decomposition, they have efficient strategies by Fourier transform (Theorem 1 with Lemma 1) and recursive formula (Theorem 2), respectively.

\begin{figure}[!t]
	\centering
	\includegraphics[scale=0.4]{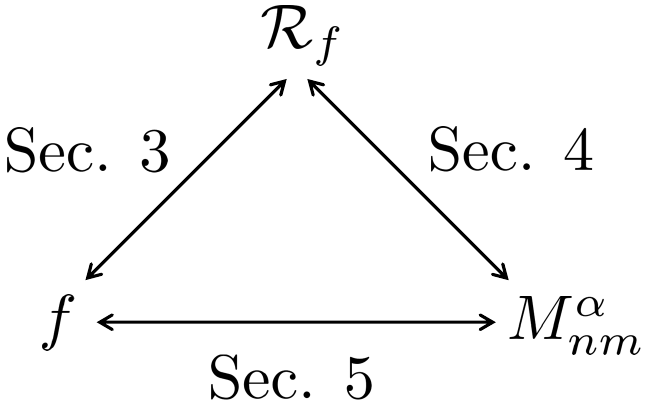}
	\centering
	\caption{Illustration for the relationship between the implicit and explicit paths for defining FMR. The explicit path enables a direct transformation from the original image $f$ to the FMR $M_{nm}^\alpha$ without the Radon transform ${{\cal R}_f}$ as an intermediary.}
\end{figure}

\section{Fractional-order Moments in Radon:\\ An Explicit Path}

In this section, unlike the above implicit path that must pass through the Radon transform, we attempts to give an explicit counterpart directly from the image to the FMR.

\subsection{From Implicit to Explicit}

The relationship between the implicit and explicit paths for defining FMR is illustrated in Fig. 5. As a broad perspective, in Section 3, the projection from the original image $f$ to the Radon representation ${{\cal R}_f}$ is formalized on the basis ${{\cal B}_1} = \delta (r - x\cos \theta  - y\sin \theta )$; in Section 4, the projection from the Radon representation ${{\cal R}_f}$ to the fractional-order orthogonal moments $M_{nm}^\alpha $ is formalized on the basis ${{\cal B}_2} = V_{nm}^\alpha (r,\theta )$. Therefore, a natural idea is to derive a \emph{direct projection} from original image $f$ to FMR $M_{nm}^\alpha $, with the cascade basis functions ${\cal B} = {{\cal B}_1} \circ {{\cal B}_2}$. 

Note that the following formal definitions are mainly provided as useful supplementary material for the future study on the behavior of FMR. In this paper, the analyses and experiments are still derived from the implicit definitions of Sections 3 and 4.

\begin{figure*}[!t]
	\centering
	\subfigure[CSIQ with Gaussian noise]{\includegraphics[scale=0.75]{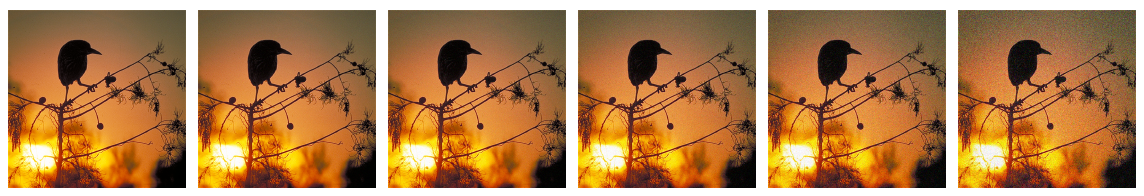}}
	\subfigure[CIDIQ with Poisson noise]{\includegraphics[scale=0.75]{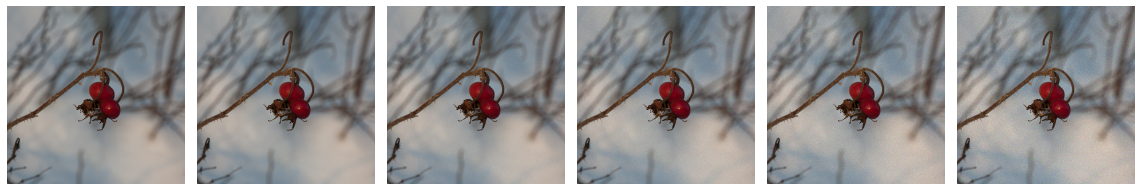}}
	\centering
	\caption{Samples of noisy images with different levels (from right to left) of Gaussian/Poisson noise from the dataset CSIQ/CIDIQ. Best viewed in color and zoom.}
\end{figure*}

\subsection{Explicit Definition}

For the FMR with harmonic radial basis function of Definition 3 (i.e., harmonic FMR), we introduce the following explicit definition, which can be interpreted as a linear combination of two integral terms.

\begin{theorem} \textbf{(Harmonic FMR: Explicit definition).}
	The harmonic FMR with Definitions 2 and 3 can be rewritten into an explicit form \cite{ref24}, as follows:
	\begin{equation}
		\begin{split}
		\left< {{\cal R}_f},V_{nm}^\alpha \right>  = \sum\limits_{k = 0}^\infty  {{W_1}^*(n,k)} \sum\limits_{t = 0}^\infty \binom{\alpha k + \frac{\alpha }{2}}{t}\\
		\times{\Theta _{\alpha k + \frac{\alpha }{2} - t,t}}{G_{\alpha k + \frac{\alpha }{2} - t,t}},
		\end{split}
	\end{equation}
	where the coefficient term ${W_1}$ is defined as:
	\begin{equation}
		{W_1}(n,k) = \frac{{\sqrt \alpha  {{(\bm{j}2n\pi )}^k}}}{{\sqrt {2\pi } k!}},
	\end{equation}
	and two integral terms $\Theta$ and $G$ are defined respectively as:
	\begin{equation}
		{\Theta _{{\xi _1},{\xi _2}}} = \int\limits_\theta  {A_m^*(\theta ){{(\cos \theta )}^{{\xi _1}}}{{(\sin \theta )}^{{\xi _2}}}d\theta },
	\end{equation}
	\begin{equation}
		{G_{{\xi _1},{\xi _2}}} = \int\limits_x {\int\limits_y {f(x,y){x^{{\xi _1}}}{y^{{\xi _2}}}dxdy} },
	\end{equation}
	where $G$ is in fact the well-known \emph{geometric moments} \cite{ref14}.
\end{theorem}

\emph{Proof.} The proof is given in Appendix B, available in the online supplemental material. \qed

\begin{theorem} \textbf{(Polynomial FMR: Explicit definition).}
	The polynomial FMR with Definitions 2 and 4 can be rewritten into an explicit form, as follows:
	\begin{equation}
		\begin{split}
			\left< {{\cal R}_f},V_{nm}^\alpha \right>  = \sum\limits_{k = 0}^n {{W_2}(\alpha ,p,q,n,k)\sum\limits_{s = 0}^\infty  {{W_3}(p,q,s)} }\\ 
			\times\sum\limits_{t = 0}^\infty \binom{\alpha (s + k + \frac{q}{2})}{t}{\Theta _{\alpha (s + k + \frac{q}{2}) - t,t}}{G_{\alpha (s + k + \frac{q}{2}) - t,t}},
		\end{split}
	\end{equation}
	where the coefficient terms ${W_2}$ and ${W_3}$ are defined as:
	\begin{equation}
		\begin{split}
		{W_2}(\alpha ,p,q,n,k) = \sqrt {\frac{{\alpha (p + 2n)  \Gamma (q + n)  n!}}{{2\pi  \Gamma (p + n) \Gamma (p - q + n + 1)}}}\\
		\times \frac{{{{( - 1)}^k}\Gamma (p + n + k)}}{{k!(n - k)!\Gamma (q + k)}},
		\end{split}
	\end{equation}
	\begin{equation}
		{W_3}(p,q,s) = {( - 1)^s}\binom{\frac{{p - q}}{2}}{s},
	\end{equation}
	with the same notations on $\Theta$ and $G$ in Theorem 3.
\end{theorem}

\emph{Proof.} The proof is given in Appendix C, available in the online supplemental material. \qed

\emph{Recalling the explicit definition.} In general, the explicit definitions in this section provide a more in-depth perspective for understanding FMR. Here, one of the valuable observations is that FMR can be interpreted as an \emph{infinite linear combination} of geometric moments, establishing a link to this very basic concept in the community. It is worth noting that, in the related work \cite{ref24}, each Radon-domain integer-order moments is a finite linear combination of geometric moments. This distinction also implies the generic nature of FMR.

\begin{figure*}[!t]
	\centering
	\subfigure[Harmonic FMR magnitude for dataset CSIQ with Gaussian noise]{\includegraphics[scale=0.19]{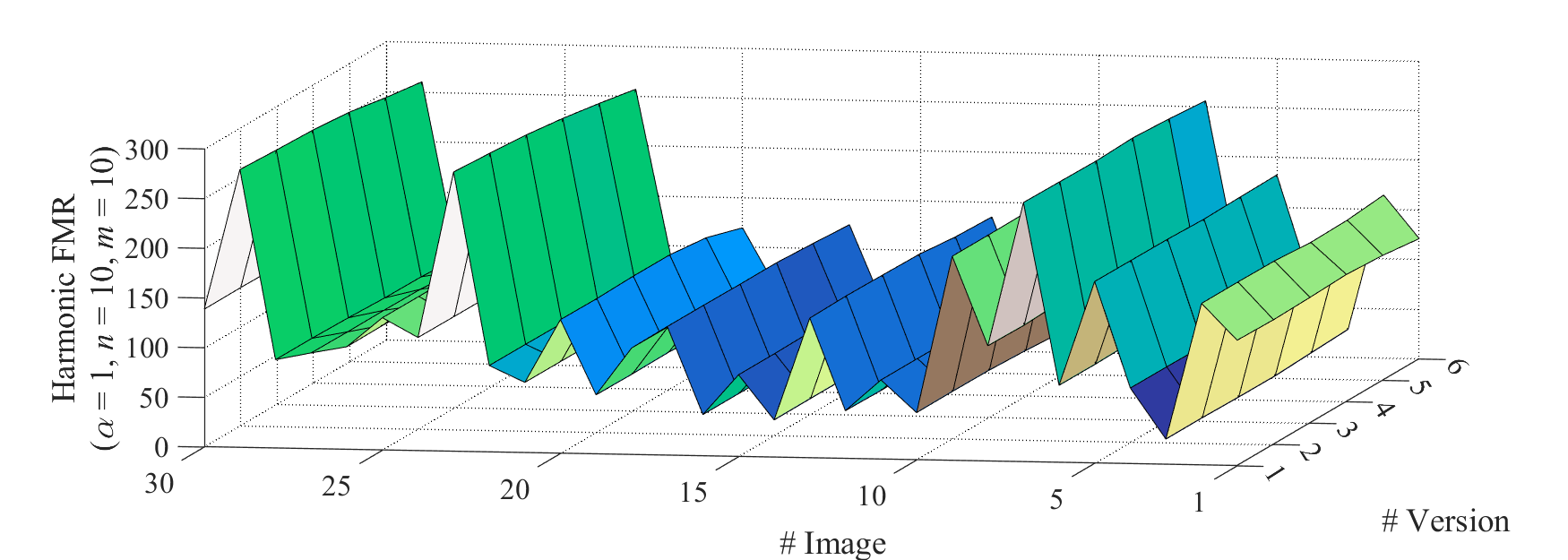}}
	\subfigure[Polynomial FMR magnitude for dataset CSIQ with Gaussian noise]{\includegraphics[scale=0.19]{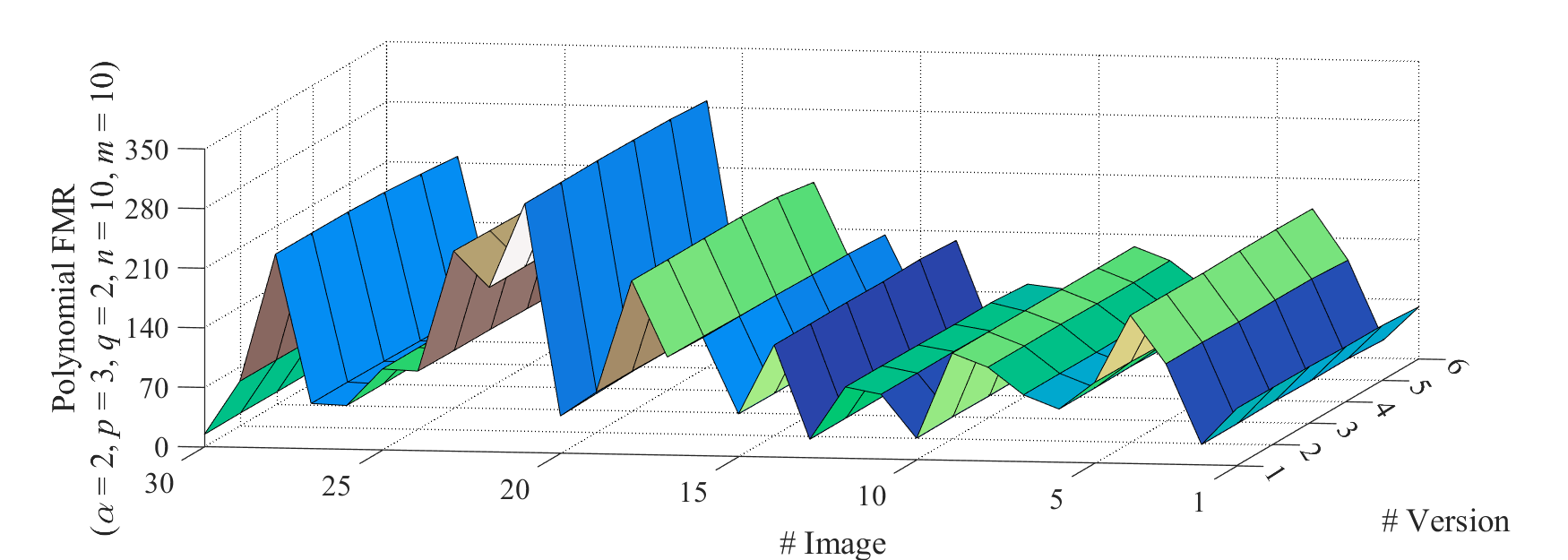}}
	\subfigure[Harmonic FMR magnitude for dataset CIDIQ with Poisson noise]{\includegraphics[scale=0.19]{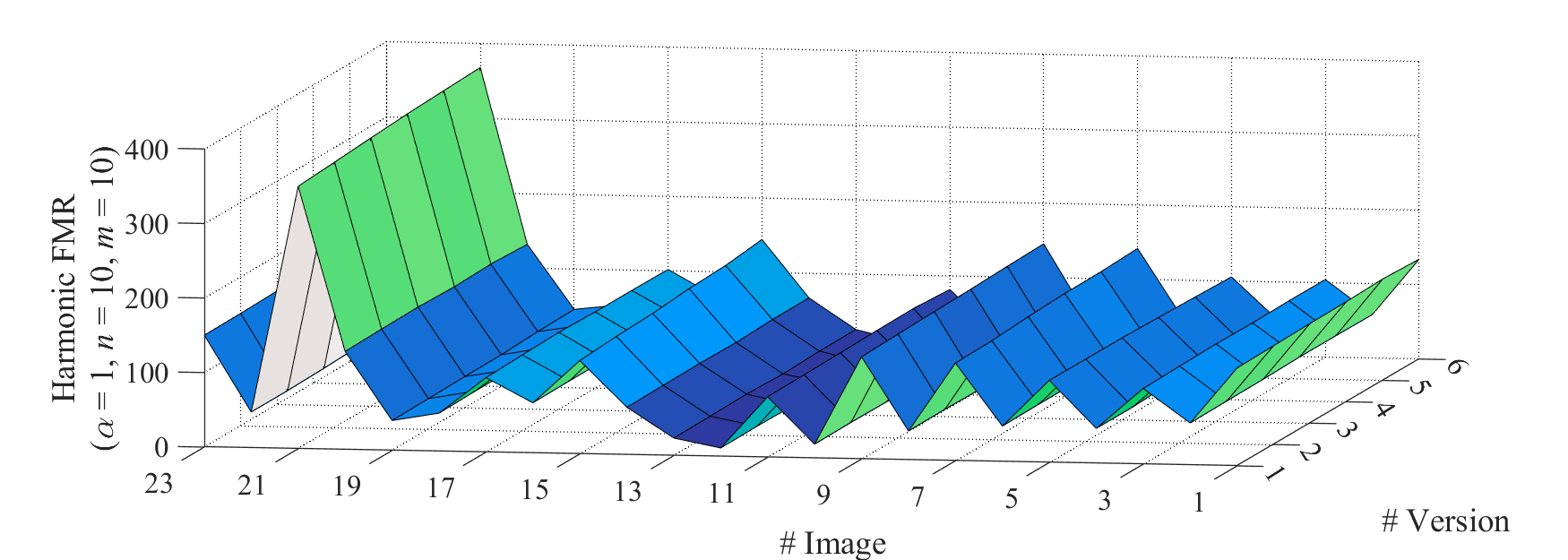}}
	\subfigure[Polynomial FMR magnitude for dataset CIDIQ with Poisson noise]{\includegraphics[scale=0.19]{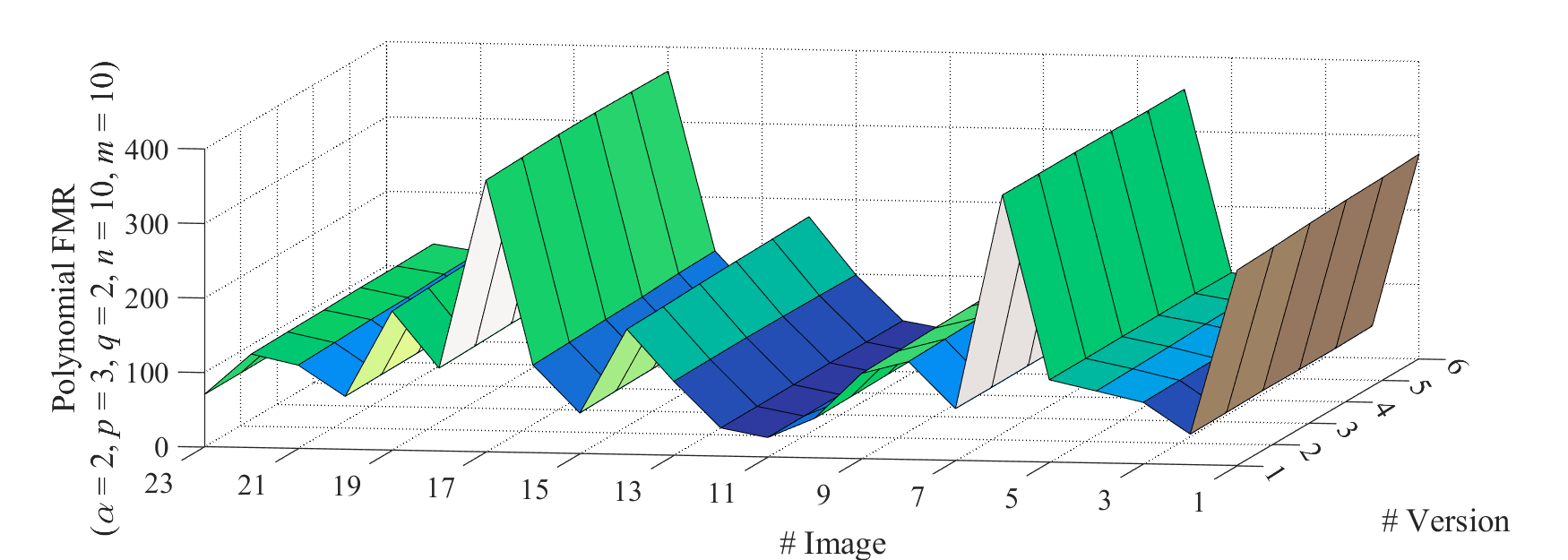}}
	\centering
	\caption{The harmonic and polynomial FMR magnitudes over different images (from right to left) with different levels of Gaussian/Poisson noise (from front to back) in dataset CSIQ/CIDIQ.}
\end{figure*}

\section{Experiments}

In this section, we will evaluate the performance of the proposed FMR, covering experiments at both the simulation level and the application level.

At the simulation level, we first visualize the histogram and reconstruction of FMR, providing a high-level perception of the noise robustness and discriminability. For a more quantitative results, we also perform a pattern recognition study on challenging noisy images, involving a comprehensive comparison with state-of-the-art learning representations (robustness enhancement via augmentation or denoising).

At the application level, we focus on the specific practice of FMR for robust visual tasks. Specifically, we first consider the challenge of template matching in the wild, for the background of trustworthy scenarios like medical image analysis. Then, we consider a zero-watermarking algorithm towards image copyright protection, for the inherent adversarial nature of security scenarios. Note that the noise robustness and geometric invariance of the image representation are crucial for achieving their core functions.

Note that all experiments are performed in Matlab R2021a, with 2.90-GHz CPU and 16-GB RAM, under Microsoft Windows environment. The code is available online at \texttt{https://github.com/ShurenQi/FMR}.

\subsection{Feature Histogram}

In this experiment, we verify the noise robustness of FMR by visualizing the histogram of a feature under different noise conditions. Also, the discriminability for different images is considered in the experiment, demonstrating the \emph{non-trivial nature} of such robust feature (i.e., not a constant value w.r.t. distinct images).

As shown in Fig. 6, the experiment is executed on the well-known degraded image datasets CSIQ \cite{ref46} and CIDIQ \cite{ref47}, involving Gaussian noise and Poisson noise, respectively. Here, the two image datasets contain 30 and 23 original images, respectively, and each original image has 5 noisy versions at different levels, resulting in a total of 318 images.

The magnitude histograms for a feature of harmonic/polynomial FMR over above clean and noisy images are presented in Fig. 7. Here, the feature of harmonic FMR is with  $\alpha  = 1$, $n = 10$, and $m = 10$, and the feature of polynomial FMR is with $\alpha  = 2$, $p = 3$, $q = 2$, $n = 10$, and $m = 10$. As shown in histograms, the feature value is always almost constant within each individual noisy series (regardless of noise level or type) while significantly different for distinct images. This common phenomenon suggests that the proposed FMR achieves ideal robustness to Gaussian and Poisson noise, while the representation is sufficiently discriminative rather than just trivially invariant.

\begin{figure*}[!t]
	\centering
	\subfigure[Original ``Fourier"]{\includegraphics[scale=0.45]{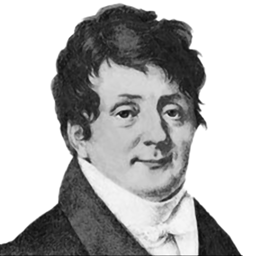}}
	\subfigure[Noisy ``Fourier"]{\includegraphics[scale=0.45]{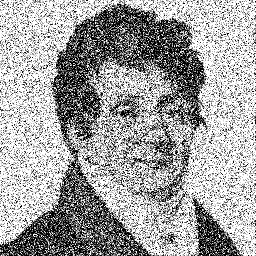}}
	\subfigure[Harmonic FM]{\includegraphics[scale=0.45]{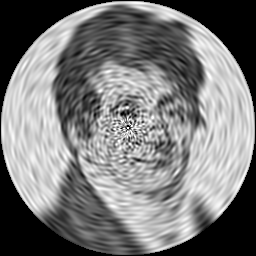}}
	\subfigure[Harmonic FMR]{\includegraphics[scale=0.45]{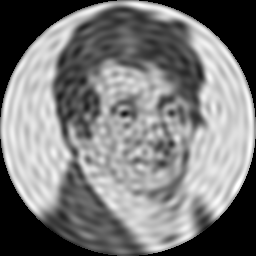}}
	
	\subfigure[Original ``Jacobi"]{\includegraphics[scale=0.45]{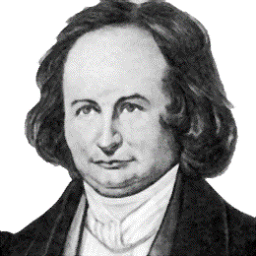}}
	\subfigure[Noisy ``Jacobi"]{\includegraphics[scale=0.45]{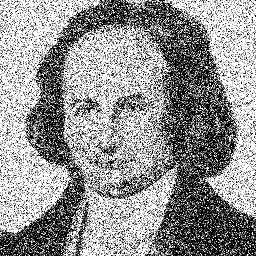}}
	\subfigure[Polynomial FM]{\includegraphics[scale=0.45]{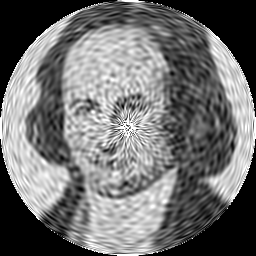}}
	\subfigure[Polynomial FMR]{\includegraphics[scale=0.45]{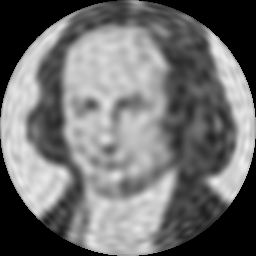}}
	\centering
	\caption{Samples of reconstruction from noisy images. (a) original image ``Fourier", (b) noisy version of ``Fourier", (c) reconstructed version by harmonic FM with MSRE = 0.060 and SSIM = 0.491 w.r.t. original ``Fourier", (d) reconstructed version by harmonic FMR with MSRE = 0.046 and SSIM = 0.596, (e) original image ``Jacobi", (f) noisy version of ``Jacobi", (g) reconstructed version by polynomial FM with MSRE = 0.062 and SSIM = 0.499 w.r.t. original ``Jacobi", (h) reconstructed version by polynomial FMR with MSRE = 0.046 and SSIM = 0.647 w.r.t. original ``Jacobi".}
\end{figure*}

\subsection{Image Reconstruction}

In this experiment, we reconstruct the noisy image from the features, visualizing what kind of information in the image is specifically captured by FMR. Starting from the reconstruction results, we intuitively know how well the features describe the image content and how much the noise interferes with the features.

Regarding the orthogonality of $V_{nm}^\alpha$, the Radon projection of image can be reconstructed from FMR $\{ M_{nm}^\alpha  =  \left< {{\cal R}_f},V_{nm}^\alpha \right> \}$ as:
\begin{equation}
	\widehat {{{\cal R}_f}} = \sum\limits_{(n,m) \in {\bf{S}}(K)} {M_{nm}^\alpha V_{nm}^\alpha }  
\end{equation}
where $\widehat {{{\cal R}_f}}$ is the reconstructed version of original Radon projection ${{\cal R}_f}$, and ${\bf{S}}(K)=\{(n,m):|n|,|m|\le K\}$ is set of
orders $(n,m)$ constrained by a integer constant $K$. After obtaining $\widehat {{{\cal R}_f}}$, the reconstructed version of the original image, $\hat f$, can also be achieved by the \emph{inverse Radon transform}. It has been well studied and is built-in to typical programming languages — technically, implemented by the Fourier transform \cite{ref48}.

\begin{figure}[!t]
	\centering
	\subfigure{\includegraphics[scale=0.75]{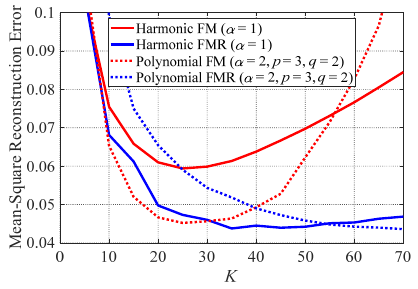}}
	\subfigure{\includegraphics[scale=0.75]{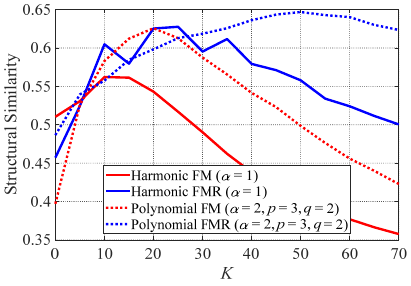}}
	\centering
	\caption{Mean-square reconstruction error and structural similarity between the reconstructed version of noise image and the original version by different methods.}
\end{figure}

The experiment is executed on the portraits of mathematicians Fourier and Jacobi in honor of the functions (7) and (8) they studied. Such images are degraded by Gaussian noise with a variance of 0.2. Note that the image-domain counterpart of FMR, i.e., the FM defined by $\left< f,V_{nm}^\alpha \right>$ \cite{ref23, ref35}, is introduced for a reasonable comparison baseline. As for the quantitative analysis, the mean-square reconstruction error (MSRE) \cite{ref49} or structural similarity (SSIM) \cite{ref50} is calculated between the original image and the reconstruction from its noisy version.

In Fig. 8, we illustrate the reconstruction from noisy images by FM and FMR. Here, harmonic FM and FMR are with $\alpha  = 1$ and $K = 30$; polynomial FM and FMR are with $\alpha  = 2$, $p = 3$, $q = 2$, and $K = 50$. Note that FMR and FM have very consistent settings here, the only difference is whether Radon projection is involved or not. Visually, the proposed FMR reconstructs the main content of clean image equally well as the FM, while reconstructing less noise than the FM. The corresponding numerical results also support this observation, where the FMR exhibits a lower MSRE and a higher SSIM compared to the FM. The above experimental results provide strong evidence for our motivation: the Radon projection does enhance noise robustness, just as predicted by the Property 3, and it is reasonable to construct our representation in such Radon space.

In Fig. 9, we provide more comprehensive quantitative results. The MSRE and SSIM curves w.r.t. $K$ show that the visual effect of FM (red lines) typically degrades significantly beyond a specific value of $K$ , where more noise being reconstructed; while our FMR (blue lines) mitigates this trend significantly, with considerable performance advantages at higher values of $K$ . These common phenomena suggest that: 1) the high-order image-domain FM is sensitive to noise, which prevents high-order moments from being used to represent images more discriminatively; 2) high-order FMR exhibits much better noise robustness, providing further quantitative evidence for Property 3.

\begin{table*}
	\renewcommand{\arraystretch}{1.3}
	\caption{Correct Classification Percentages (\%) for Classical Moment Representations, FM Representations, and FMR Representations.}
	\centering
	\begin{tabular}{ccccccccccccccc}
		\thickhline
		{\multirow{2}[1]{*}{Method}} & \multicolumn{7}{c}{$K=10$ with Gaussian white noise (variance)} & \multicolumn{7}{c}{$K=20$ with Gaussian white noise (variance)}\\
		\cmidrule(lr){2-8}\cmidrule(lr){9-15}\multicolumn{1}{c}{}    & 0     & 0.05  & 0.1   & 0.15  & 0.2   & 0.25  & 0.3      & 0     & 0.05  & 0.1   & 0.15  & 0.2   & 0.25  & 0.3 \\
		\midrule
		ART       & \textcolor{red}{\textbf{100}} & \textcolor{blue}{99.17} & \textcolor{blue}{88} & \textcolor{blue}{74.53} & \textcolor{blue}{50.14} & \textcolor{blue}{30.14} & \textcolor{blue}{19.19}     & \textcolor{red}{\textbf{100}} & \textcolor{blue}{99.39} & \textcolor{blue}{93.28} & \textcolor{blue}{80.14} & \textcolor{blue}{59.22} & \textcolor{blue}{37.42} & \textcolor{blue}{23.67}\\
		GFD       & \textcolor{red}{\textbf{100}} & \textcolor{blue}{99.28} & \textcolor{blue}{89.86} & \textcolor{blue}{74.75} & \textcolor{blue}{50.14} & \textcolor{blue}{28.67} & \textcolor{blue}{16.86}     & \textcolor{red}{\textbf{100}} & \textcolor{blue}{99.28} & \textcolor{blue}{93.69} & \textcolor{blue}{80.03} & \textcolor{blue}{60.67} & \textcolor{blue}{38.67} & \textcolor{blue}{23.53} \\
		ZM        & \textcolor{red}{\textbf{100}} & \textcolor{blue}{88.83} & \textcolor{blue}{57.19} & \textcolor{blue}{35.17} & \textcolor{blue}{25.92} & \textcolor{blue}{18.64} & \textcolor{blue}{12.86}     & \textcolor{red}{\textbf{100}} & \textcolor{blue}{90.36} & \textcolor{blue}{61.97} & \textcolor{blue}{37.83} & \textcolor{blue}{26.67} & \textcolor{blue}{19.64} & \textcolor{blue}{13.53}\\
		\hline
		OFMM/FJFM(2,2,1)     & \textcolor{red}{\textbf{100}} & \textcolor{red}{\textbf{100}} & \textcolor{blue}{95.28} & \textcolor{blue}{82.94} & \textcolor{blue}{60.89} & \textcolor{blue}{36.03} & \textcolor{blue}{22.36}     & \textcolor{red}{\textbf{100}} & \textcolor{red}{\textbf{100}} & 97.06 & \textcolor{blue}{88.44} & \textcolor{blue}{71.33} & \textcolor{blue}{48.22} & \textcolor{blue}{30.31}\\
		CHFM/FJFM(2,1.5,1)     & \textcolor{red}{\textbf{100}} & \textcolor{red}{\textbf{100}} & 97.78 & 86.78 & 70.97 & 51.19 & \textcolor{red}{36.22}     & \textcolor{red}{\textbf{100}} & \textcolor{red}{\textbf{100}} & 98    & 91.81 & 80.14 & 59.14 & 45.31 \\
		JFM(3.3)/FJFM(3,3,1)     & \textcolor{red}{\textbf{100}} & \textcolor{red}{\textbf{100}} & 97.08 & 85.92 & 71.92 & 47.31 & 30.61     & \textcolor{red}{\textbf{100}} & \textcolor{red}{\textbf{100}} & 97.61 & 90.78 & 78.39 & 57.83 & 36.86 \\
		FJFM(3,3,2)     & \textcolor{red}{\textbf{100}} & \textcolor{red}{\textbf{100}} & 98.56 & 86.11 & 75.92 & \textcolor{red}{51.86} & \textcolor{red}{\textbf{36.75}}     & \textcolor{red}{\textbf{100}} & \textcolor{red}{\textbf{100}} & 98.64 & 91.19 & 82.81 & 63.89 & 44.83 \\
		EFM/GPCET(1)     & \textcolor{red}{\textbf{100}} & \textcolor{red}{\textbf{100}} & 97.28 & 90.47 & 68.25 & 46.39 & 28.83     & \textcolor{red}{\textbf{100}} & \textcolor{red}{\textbf{100}} & 97.08 & 91.28 & 78.06 & 56.42 & 40.56 \\
		PCET/GPCET(2)     & \textcolor{red}{\textbf{100}} & \textcolor{red}{\textbf{100}} & \textcolor{blue}{95.36} & \textcolor{blue}{83.33} & \textcolor{blue}{60.17} & \textcolor{blue}{35.5} & \textcolor{blue}{22.06}    & \textcolor{red}{\textbf{100}} & \textcolor{red}{\textbf{100}} & \textcolor{blue}{96} & \textcolor{blue}{87.61} & \textcolor{blue}{71.06} & \textcolor{blue}{45.92} & \textcolor{blue}{29.69} \\
		GPCET(1.5)     & \textcolor{red}{\textbf{100}} & \textcolor{red}{\textbf{100}} & 96.92 & 84.78 & 67    & 43.72 & 27.72     & \textcolor{red}{\textbf{100}} & \textcolor{red}{\textbf{100}} & \textcolor{blue}{97} & 88.94 & 75.89 & 52.64 & 34.17\\
		\hline
		\rowcolor{mygray2}OFMMR/FJFMR(2,2,1)     & \textcolor{red}{\textbf{100}} & \textcolor{red}{\textbf{100}} & \textcolor{red}{98.89} & \textcolor{red}{91.69} & \textcolor{red}{77.56} & \textcolor{red}{52.89} & 32.97     & \textcolor{red}{\textbf{100}} & \textcolor{red}{\textbf{100}} & \textcolor{red}{\textbf{100}} & \textcolor{red}{97.89} & \textcolor{red}{95.53} & \textcolor{red}{84.89} & \textcolor{red}{57.47}\\
		\rowcolor{mygray2}CHFMR/FJFMR(2,1.5,1)     & \textcolor{red}{\textbf{100}} & \textcolor{red}{\textbf{100}} & 98.11 & \textcolor{red}{92.03} & \textcolor{red}{77.75} & \textcolor{red}{53.31} & 32.75     & \textcolor{red}{\textbf{100}} & \textcolor{red}{\textbf{100}} & \textcolor{red}{\textbf{100}} & \textcolor{red}{98.06} & \textcolor{red}{95.72} & \textcolor{red}{\textbf{85.53}} & \textcolor{red}{57.19} \\
		\rowcolor{mygray2}JFMR(3.3)/FJFMR(3,3,1)     & \textcolor{red}{\textbf{100}} & \textcolor{red}{\textbf{100}} & \textcolor{red}{\textbf{99}} & \textcolor{red}{\textbf{93.39}} & \textcolor{red}{78.89} & \textcolor{red}{54.42} & \textcolor{red}{33.39}     & \textcolor{red}{\textbf{100}} & \textcolor{red}{\textbf{100}} & \textcolor{red}{\textbf{100}} & \textcolor{red}{\textbf{98.33}} & \textcolor{red}{\textbf{96.42}} & \textcolor{red}{84.5} & \textcolor{red}{57.42} \\
		\rowcolor{mygray2}FJFMR(3,3,2)     & \textcolor{red}{\textbf{100}} & \textcolor{red}{\textbf{100}} & 98.61 & 86.19 & 71.67 & 47.47 & \textcolor{red}{33}     & \textcolor{red}{\textbf{100}} & \textcolor{red}{\textbf{100}} & 99.67 & 98    & 90    & \textcolor{red}{78.17} & 52.28 \\
		\rowcolor{mygray2}EFMR/GPCETR(1)     & \textcolor{red}{\textbf{100}} & \textcolor{red}{\textbf{100}} & \textcolor{red}{\textbf{99}} & \textcolor{red}{92.56} & \textcolor{red}{\textbf{82.36}} & \textcolor{red}{\textbf{56.47}} & \textcolor{red}{36.47}     & \textcolor{red}{\textbf{100}} & \textcolor{red}{\textbf{100}} & \textcolor{red}{99.06} & \textcolor{red}{97.14} & \textcolor{red}{92.72} & \textcolor{red}{79.78} & \textcolor{red}{\textbf{58.5}} \\
		\rowcolor{mygray2}PCETR/GPCETR(2)     & \textcolor{red}{\textbf{100}} & \textcolor{red}{\textbf{100}} & \textcolor{red}{98.94} & 89.36 & 73.53 & 48.08 & 31.03     & \textcolor{red}{\textbf{100}} & \textcolor{red}{\textbf{100}} & 99.06 & 95.08 & 88.94 & 72.14 & \textcolor{red}{52.44} \\
		\rowcolor{mygray2}GPCETR(1.5)     & \textcolor{red}{\textbf{100}} & \textcolor{red}{\textbf{100}} & \textcolor{red}{98.83} & \textcolor{red}{92.61} & \textcolor{red}{78.44} & 50.42 & 31.78     & \textcolor{red}{\textbf{100}} & \textcolor{red}{\textbf{100}} & 98.92 & 95.42 & \textcolor{red}{90.78} & 75.64 & 52\\
		\bottomrule
	\end{tabular}%
\end{table*}

\begin{figure}[!t]
	\centering
	\subfigure{\includegraphics[scale=0.36]{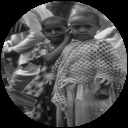}}
	\subfigure{\includegraphics[scale=0.36]{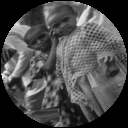}}
	\subfigure{\includegraphics[scale=0.36]{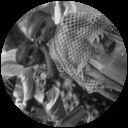}}
	\subfigure{\includegraphics[scale=0.36]{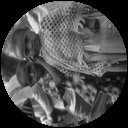}}
	\subfigure{\includegraphics[scale=0.36]{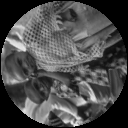}}
	\subfigure{\includegraphics[scale=0.36]{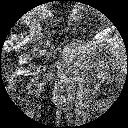}}
	\subfigure{\includegraphics[scale=0.36]{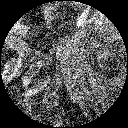}}
	\subfigure{\includegraphics[scale=0.36]{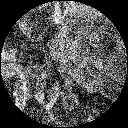}}
	\subfigure{\includegraphics[scale=0.36]{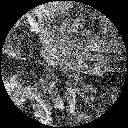}}
	\subfigure{\includegraphics[scale=0.36]{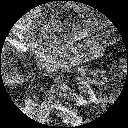}}
	\subfigure{\includegraphics[scale=0.36]{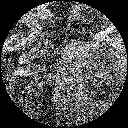}}
	\subfigure{\includegraphics[scale=0.36]{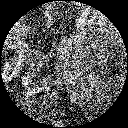}}
	\subfigure{\includegraphics[scale=0.36]{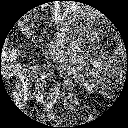}}
	\subfigure{\includegraphics[scale=0.36]{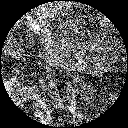}}
	\subfigure{\includegraphics[scale=0.36]{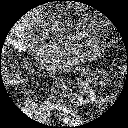}}
	\centering
	\caption{Some samples of the testing images with rotation angles $\{ 0^\circ ,30^\circ ,...,120^\circ \} $ (from left to right) and noise variances $\{ 0,0.05,0.1\} $ (from top to bottom).}
\end{figure}
\subsection{Pattern Recognition}

In this experiment, we perform the pattern recognition on the images with noise and orientation variations, directly verifying the rotation invariance, noise robustness, and discriminative power of FMR. Here, we provide a detailed score comparison with the image-domain counterpart of FMR, i.e., the FM, at multiple parameter settings. The score comparison also covers some learning representations, and their robustness is enhanced through augmentation and/or denoising. Such comparisons position the proposed approach w.r.t. the current state of the art in small-scale robust vision scenarios.

The experimental images are derived from the COREL dataset \cite{ref51}, where 100 images are selected and normalized to a size of $128 \times 128$ as original images. For the non-learning FM and FMR, the classifier will be trained only on the corresponding representations of these original images. While the testing will be performed on the degraded versions of the original images, the degradation operation includes Gaussian noise with variances $\{ 0,0.05,...,0.3\} $ and rotation with angles $\{ 0^\circ ,10^\circ ,...,350^\circ \} $, as shown in Fig. 10. Therefore, the testing set contains 25200 images. As for the learning representations, only one training image for a category is generally not enough to form a reliable representation, so rotation (36 samples) or/and noise (36 samples) augmented versions are also included in the training. Obviously, such augmentation will result in a 36-fold or 72-fold increase in training size w.r.t. the non-learning FM and FMR. In addition to above noise augmentation, the denoising preprocessing of testing image is also used for the learning representations, as another means of robustness enhancement. Note that the naive minimum-distance classifier is adopted here for reflecting the nature of the representations themselves. In addition, the denoising will be implemented by a promising method with deep neural network \cite{ref52}.

\begin{table*}
	\renewcommand{\arraystretch}{1.3}
	\caption{Correct Classification Percentages (\%) for Learning Methods (With Augmentation and Denoising) Compared to the Average/Best Results of FM and FMR Representations.}
	\centering
	\begin{tabular}{cccccccccc}
		\hline
		\multirow{2}[1]{*}{Method} & \multirow{2}[1]{*}{\tabincell{c}{Train.\\size}} & \multirow{2}[1]{*}{\tabincell{c}{Pre\\process.}} & \multicolumn{7}{c}{Gaussian white noise (variance)} \\
		\cmidrule(lr){4-10}\multicolumn{1}{c}{} &  &  & 0     & 0.05  & 0.1   & 0.15  & 0.2   & 0.25  & 0.3 \\
		\midrule
		PCANet (naive) & $\times$1    &     & 19.64 & 12.31 & 9.06  & 6.72  & 5.31  & 4.56  & 4.44\\
		PCANet (rotation augmentation) & $\times$36   &     & \textcolor{red}{100} & 51.94 & 23.22 & 11.67 & 7.64  & 5.03  & 4.69 \\
		PCANet (rotation and noise augmentation) & $\times$48   &     & \textcolor{red}{100} & 38.25 & 13.72 & 6.25  & 3.72  & 2.58  & 1.97 \\
		PCANet (rotation augmentation and denoising) & $\times$36   & \checkmark & 97    & 68.53 & 39.08 & 24.19 & 16.19 & 12.75 & 9.14\\
		\hline
		CBFD (naive) & $\times$1    &     & 24.94 & 4.27  & 1.92  & 1.81  & 1.28  & 1.17  & 0.92\\
		CBFD (rotation augmentation) & $\times$36   &     & \textcolor{red}{100} & 4.53  & 2.56  & 1.97  & 1.53  & 1.47  & 1.78 \\
		CBFD (rotation and noise augmentation) & $\times$48   &     & \textcolor{red}{100} & 11.97 & 4.61  & 2     & 1.56  & 1.53  & 1.42 \\
		CBFD (rotation augmentation and denoising) & $\times$36   & \checkmark & 90.5  & 15.75 & 9.08  & 5.58  & 4.17  & 3.33  & 2.67 \\
		\hline
		GoogLeNet (rotation augmentation) & $\times$36   &     & 99.97 & 6.08  & 2.22  & 1.89  & 1.42  & 1.28  & 1.33 \\
		GoogLeNet (rotation and noise augmentation) & $\times$72   &     & 99.97 & 69.19 & 52.47 & 39.22 & 28.25 & 19.58 & 13.14 \\
		GoogLeNet (rotation augmentation and denoising) & $\times$36   & \checkmark & \textcolor{red}{100} & 44.69 & 19.81 & 8.42  & 4.28  & 3.28  & 2.75\\
		\hline
		ResNet-50 (rotation augmentation) & $\times$36   &     & \textcolor{red}{100} & 3.44  & 1.28  & 1.03  & 1.03  & 1.19  & 1.25\\
		ResNet-50 (rotation and noise augmentation) & $\times$72   &     & \textcolor{red}{100} & 78.06 & 66.33 & 59.58 & 54.22 & 49.19 & 44.47 \\
		ResNet-50 (rotation augmentation and denoising) & $\times$36   & \checkmark & \textcolor{red}{100} & 51.81 & 23.78 & 11.47 & 6.44  & 4.86  & 4\\
		\hline
		FM (average results) & $\times$1    &     & \textcolor{red}{100} & \textcolor{red}{100} & 97.12 & 87.88 & 72.34 & 49.72 & 33.31 \\
		FM (best results) & $\times$1    &     & \textcolor{red}{100}   & \textcolor{red}{100}   & 98.64 & 91.81 & 82.81 & 63.89 & 45.31 \\
		\hline
		\rowcolor{mygray2}FMR (average results) & $\times$1    &     & \textcolor{red}{100} & \textcolor{red}{100} & 99.15 & 94.13 & 85.02 & 65.98 & 44.19 \\
		\rowcolor{mygray2}FMR (best results)& $\times$1    &     & \textcolor{red}{100}   & \textcolor{red}{100}   & \textcolor{red}{100}   & \textcolor{red}{98.33} & \textcolor{red}{96.42} & \textcolor{red}{85.53} & \textcolor{red}{58.5}\\
		\bottomrule
	\end{tabular}%
	
\end{table*}

The specific methods involved in this experiment are

\begin{itemize}
	\item	The classical moment representations (early): ART \cite{ref53}, GFD \cite{ref54}, and ZM \cite{ref16};
	\item	The FM representations (recent): FJFM ($p$, $q$, $\alpha$) \cite{ref23} and GPCET ($\alpha$) \cite{ref35}, where some well-known integer-order moments, OFMM \cite{ref17}, CHFM \cite{ref18}, JFM \cite{ref19}, EFM \cite{ref20}, and PCET \cite{ref21}, are covered by taking specific values of the parameters;
	\item	The FMR representations (ours): all the Radon-domain counterparts for above fractional-order moments;
	\item	The lightweight learning representations (recent): PCANet \cite{ref55} and CBFD \cite{ref56} with rotation/noise augmentation or denoising preprocessing \cite{ref52};
	\item	The deep learning representations (recent): GoogLeNet \cite{ref57} and ResNet-50 \cite{ref58} with rotation/ noise augmentation or denoising preprocessing \cite{ref52}.
\end{itemize}

For above moment-based representations, the magnitudes of the calculated moments on the set ${\bf{S}}(K)=\{(n,m):|n|,|m|\le K\}$ are treated as the feature vector. Since the high-order FMR is much more robust to noise than the image-domain counterpart (see also Section 6.2), we introduce a factor $(n + 1)$ on the feature vector of FMR to balance the features and enhance the discriminability. Note that this strategy is potentially harmful for image-domain moment representations and is therefore not used. The lightweight learning representations follow the typical settings in their original papers. The deep learning representations achieve the adaptability to experimental images in a transfer learning manner.

\begin{figure*}[!t]
	\centering
	\subfigure[Temp.]{\includegraphics[scale=0.1]{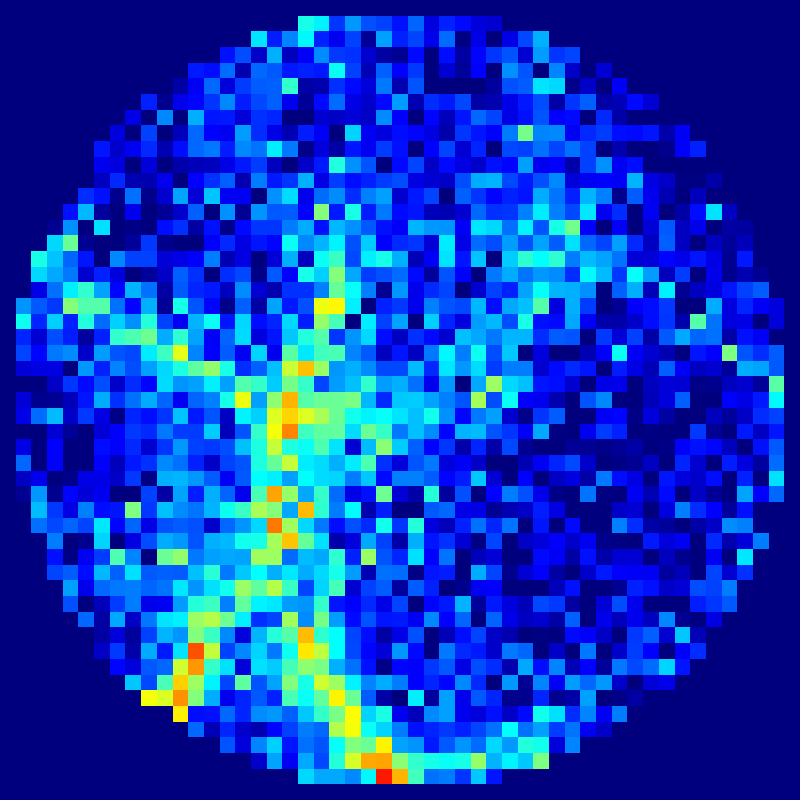}}
	\subfigure[Microscopy]{\includegraphics[scale=0.25]{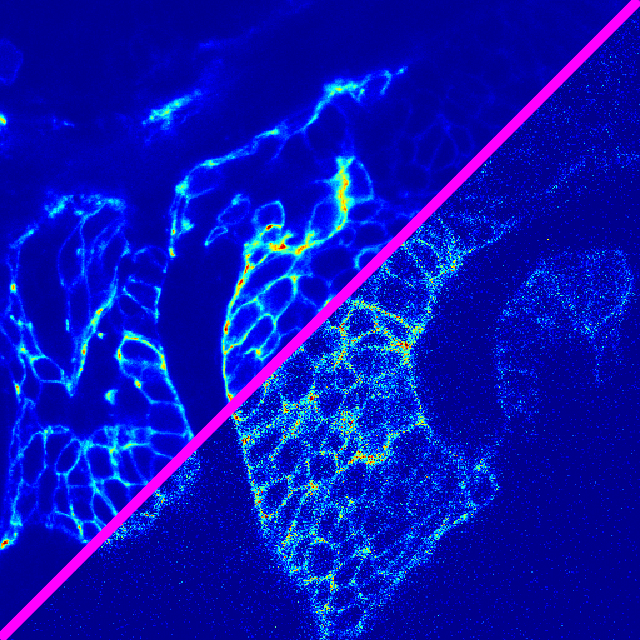}}
	\subfigure[SID]{\includegraphics[scale=0.25]{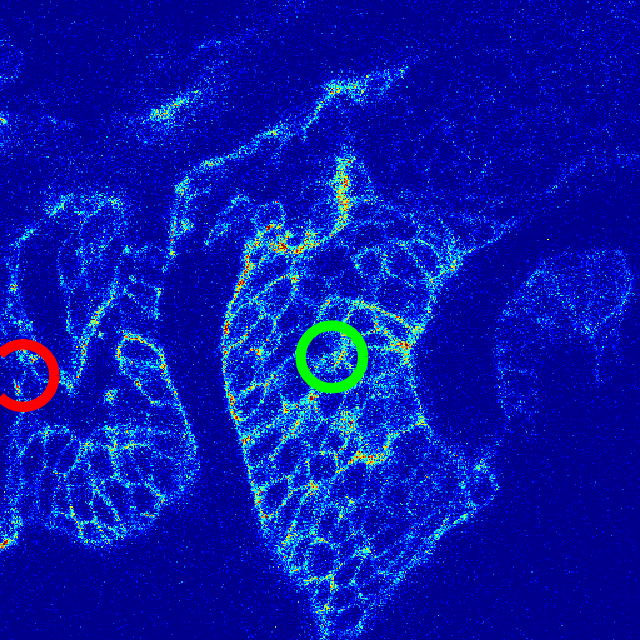}}
	\subfigure[DAISY]{\includegraphics[scale=0.25]{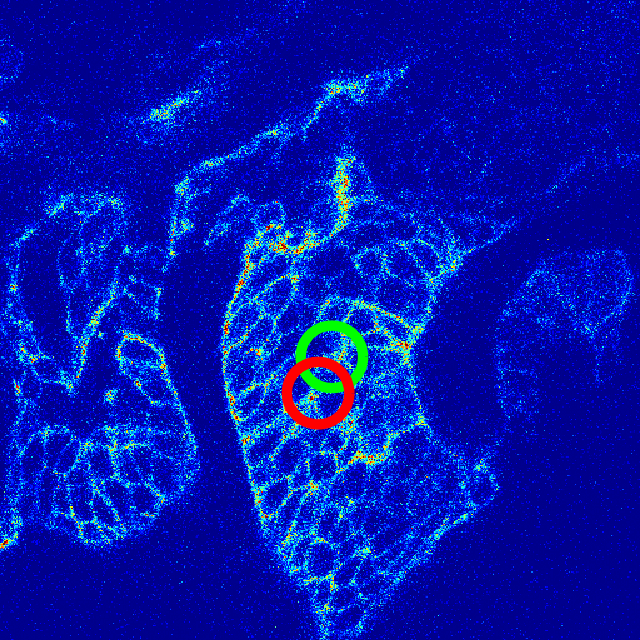}}
	\subfigure[DASC]{\includegraphics[scale=0.25]{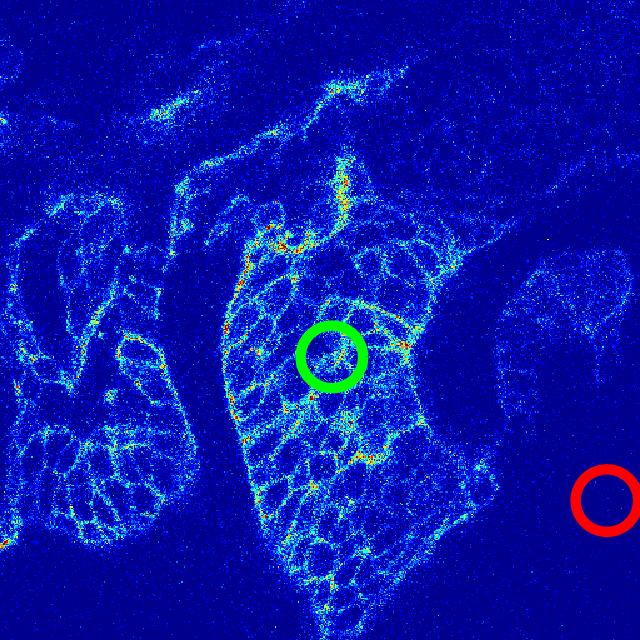}}
	\subfigure[DDIS]{\includegraphics[scale=0.25]{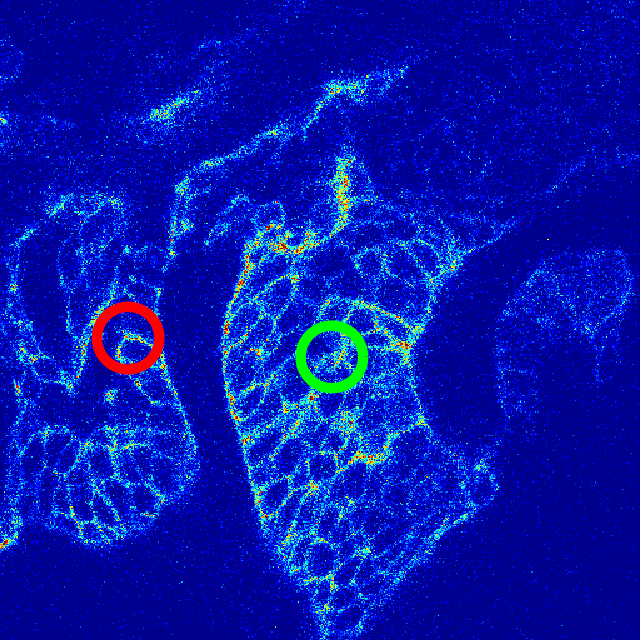}}
	\subfigure[FMR]{\includegraphics[scale=0.25]{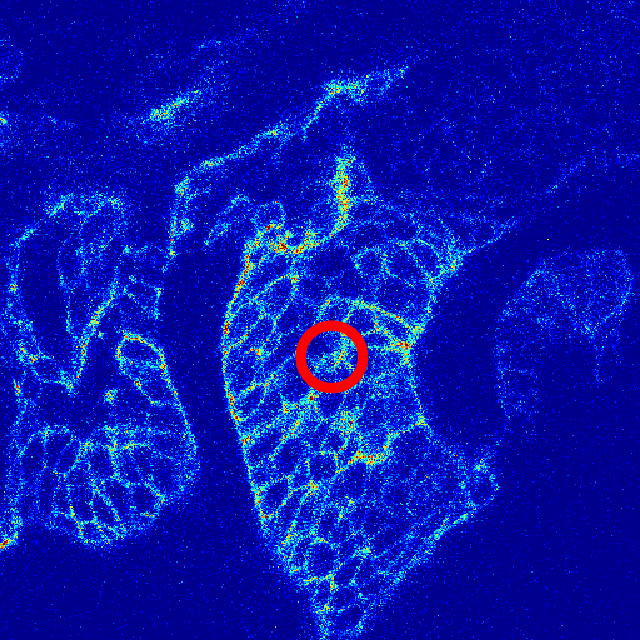}}

	\subfigure[Temp.]{\includegraphics[scale=0.1]{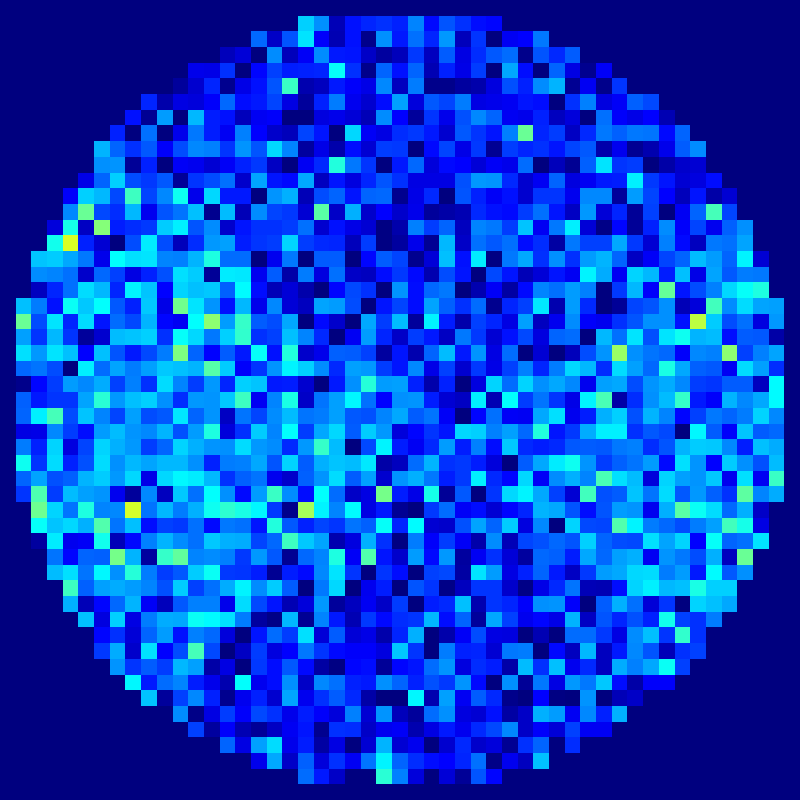}}
	\subfigure[Smartphone]{\includegraphics[scale=0.25]{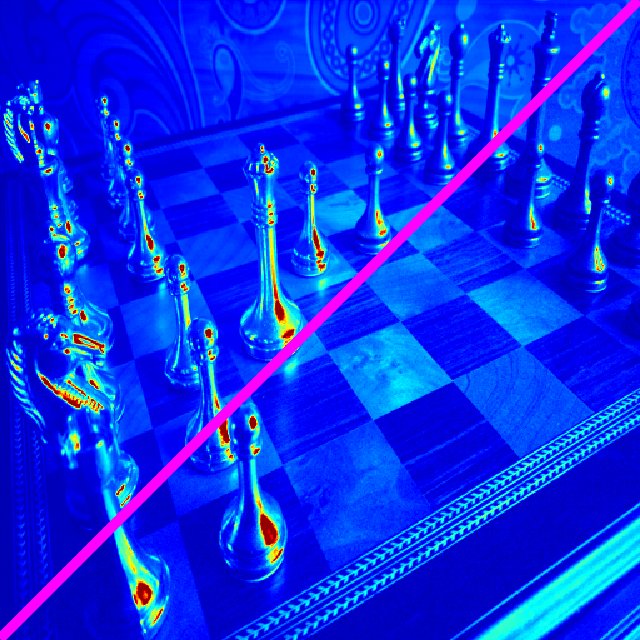}}
	\subfigure[SID]{\includegraphics[scale=0.25]{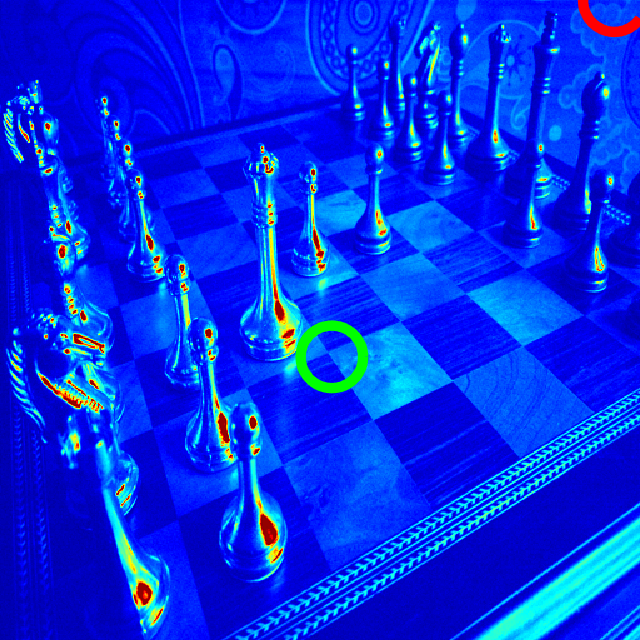}}
	\subfigure[DAISY]{\includegraphics[scale=0.25]{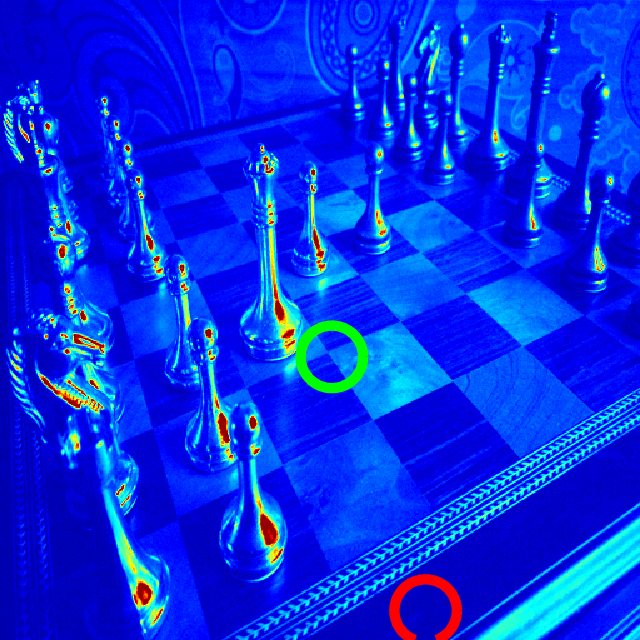}}
	\subfigure[DASC]{\includegraphics[scale=0.25]{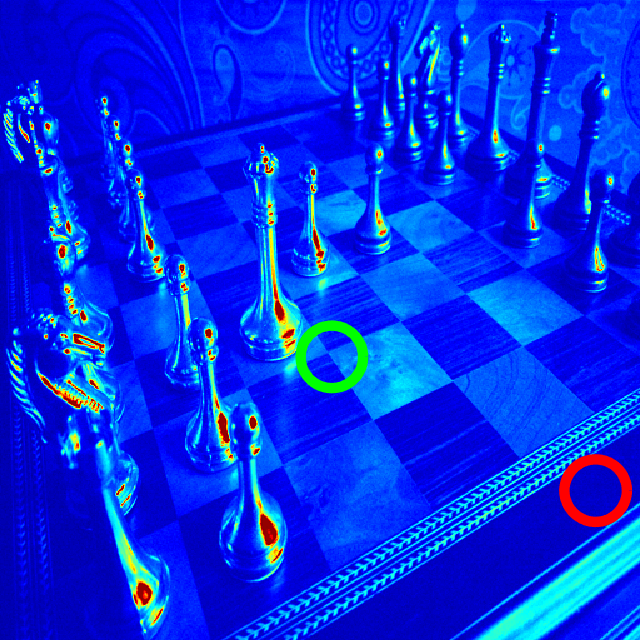}}
	\subfigure[DDIS]{\includegraphics[scale=0.25]{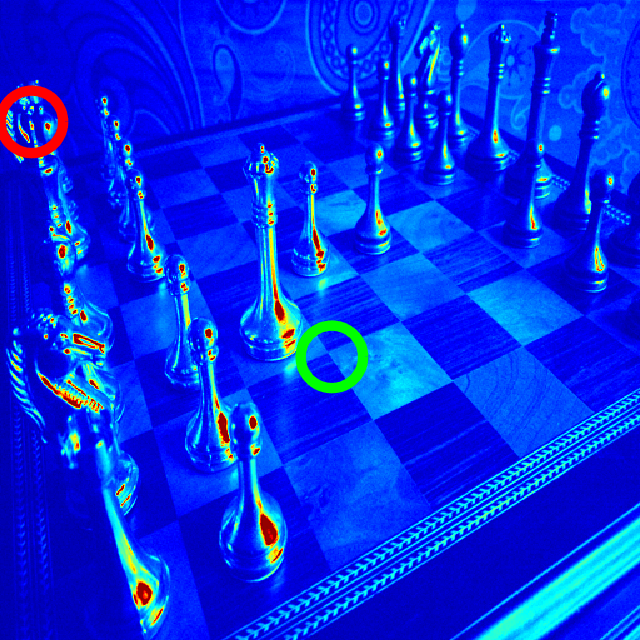}}
	\subfigure[FMR]{\includegraphics[scale=0.25]{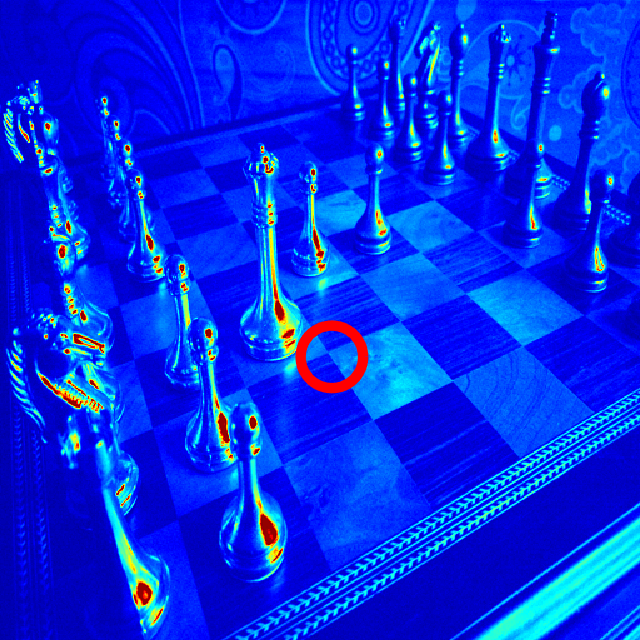}}
	
	\centering
	\caption{Samples of template matching in the wild. Here, (a) and (h) are arbitrarily oriented templates under noisy environments; (b) and (i) are images captured by fluorescence microscopy and smartphone with real imaging noise -- clean (top left) vs. noisy (bottom right), best viewed in color and zoom; (c)$\sim$(g) and (j)$\sim$(n) are results by different representations with matched (red) and ground-truth (green) targets.}
\end{figure*}

In Table 2, we list the correct classification percentages for above moment-based representations with $K = \{ 10,20\}$, where the top/bottom ranked performance is highlighted in red/blue, and the best performance highlighted in bold.

\begin{itemize}
	\item For the early and classical moment representations, their performance is typically poor in comparison to other competing methods. The possible reason is the low discriminability resulting from the non-orthogonal/non-complete basis functions.
	\item The FM provides relatively consistent performance under different parameter settings, outperforming the classical methods especially when the noise is strong.
	\item In contrast, our FMR achieves performance gains w.r.t. its FM counterpart at almost all parameter settings and all noise variances under two values of $K$. With such common gains, the FMR exhibits an overall best performance beyond the classical moment representations and FM representations. 
\end{itemize}

In general, the above results provide numerical evidence for rotation invariance, noise robustness, and discriminative power of FMR, demonstrating the benefits w.r.t. these closely related works.

In Table 3, we also list the correct classification percentages for learning representations, along with average/best results of FM and FMR representations in Table 2. Note that we complement more comprehensive results on the data augmentation and denoising preprocessing in Appendix D.

\begin{itemize}
	\item As can be observed, both lightweight learning representations fail completely, even on clean testing images, when using the same training images of moment-based representations. The main cause is that PCANet and CBFD do not have in-form rotation invariance. 
	\item After introducing rotation augmentation in training, all learning representations achieve $\sim$ 100\% correct classification on clean testing images, but at the cost of a 36-fold increase in training size. On the other hand, their performance drops sharply with increasing noise variance, implying a lack of noise robustness.
	\item When both rotation and noise augmentations are introduced, there is a significant improvement in the robustness of the two deep learning representations, while the lightweight learning representations do not exhibit this phenomenon probably due to the difference in model capacity. Note that this improvement in robustness comes at the cost of a 72-fold increase in training size — a remarkable efficiency degradation.
	\item As another way to improve noise robustness, the denoising preprocessing is also introduced here, but the resulting performance gain seems to be less pronounced than generally expected. Clearly, there is still a gap between image restoration and image recognition, where the additional denoising artifacts strongly interferes with learning representation and subsequent recognition. We note that this gap was also pointed out in very latest research \cite{ref12, ref13}.
	\item In contrast, the FM and FMR perform significantly better in terms of rotation invariance, noise robustness, and computational efficiency, even after averaging the results. Furthermore, the proposed FMR achieves an overall best performance beyond the moment-based representations and lightweight/deep learning representations. 
\end{itemize}

The above results suggest that, for small-scale robust vision problems, the proposed FMR is highly competitive even in the deep-learning era.

\subsection{Template Matching}

\begin{table}
	\renewcommand{\arraystretch}{1.3}
	\caption{Template Matching Accuracy (\%) for Different Representations in Real Fluorescence Microscopy Images.}
	\centering
	\setlength{\tabcolsep}{3mm}{
		\begin{tabular}{ccccc}
			\hline
			Method & Ideal & Rotation & Noise  & Rot.\&Noi. \\
			\midrule
			SID   & \textcolor{blue}{50}    & 38    & \textcolor{blue}{31}    & 20 \\
			DAISY & 73    & \textcolor{blue}{11}    & 50    & \textcolor{blue}{11} \\
			DASC  & 80    & \textcolor{blue}{7}     & \textcolor{blue}{22}    & \textcolor{blue}{2} \\
			DDIS  & \textcolor{blue}{68}    & 67    & 65    & 65 \\
			\rowcolor{mygray2}FMR   & \textcolor{red}{99}    & \textcolor{red}{99}    & \textcolor{red}{96}    & \textcolor{red}{93} \\
			\bottomrule
	\end{tabular} }
\end{table}

\begin{table}
	\renewcommand{\arraystretch}{1.3}
	\caption{Template Matching Accuracy (\%) for Different Representations in Real Smartphone Images.}
	\centering
	\setlength{\tabcolsep}{3mm}{
		\begin{tabular}{ccccc}
			\hline
			Method & Ideal & Rotation & Noise  & Rot.\&Noi. \\
			\midrule
			SID   & \textcolor{blue}{81}    & 64    & 27    & 20 \\
			DAISY & 99    & \textcolor{blue}{13}    & \textcolor{blue}{7}     & \textcolor{blue}{0} \\
			DASC  & 99    & \textcolor{blue}{10}    & \textcolor{blue}{5}     & \textcolor{blue}{1} \\
			DDIS  & \textcolor{blue}{71}    & 71    & 71    & 70 \\
			\rowcolor{mygray2}FMR   & \textcolor{red}{100}   & \textcolor{red}{99}    & \textcolor{red}{89}    & \textcolor{red}{83} \\
			\bottomrule
	\end{tabular} }
\end{table}

Template matching \cite{ref78}, i.e., seeking a given template in an image, is an important low-level vision task with fundamental applications in numerous image processing problems, as well as trustworthy scenarios like medical image analysis.

As a major difficulty, various (signal and geometric) inconsistencies between the template and the target typically lead to robustness issues. In this paper, we consider improving such template matching \emph{in the wild} by a direct application of FMR, covering a comparison with state-of-the-art solutions.

As shown in Fig. 11, the experiments are performed on real images captured by fluorescence microscopes \cite{ref79} and smartphones \cite{ref80}, respectively. Note that such images typically exhibit severe imaging noise, due to the inherent limitations of lighting environment and imaging hardware. In addition, experiment images exhibit strong non-local self-similarity, e.g., similar cells, which also leads to discriminative challenges. From a practical perspective of arbitrarily oriented template matching under noisy environments, we consider the following four classes of template variants:
\begin{itemize}
	\item	Ideal: well imaged and aligned templates; 
	\item	Rotation: well imaged but unaligned templates;
	\item	Noise: poorly imaged but aligned templates;
	\item	Rotation and Noise: poorly imaged and unaligned templates.
\end{itemize}

Ideal templates are formed by randomly selecting 20 regions in each clean image, with 30 pairs of clean and noisy images, resulting in a total of 600 ideal templates. Further, template variants are formed by applying arbitrary angular rotation and/or Gaussian noise with a variance of 0.01, resulting a total of 2400 templates in the experiment. Technically, the global representation of template will be compared with the local representation of analysis image (with $512\times512$ pixels), hence matching the counterpart of the template with the minimum distance. Therefore, the size of this problem is each of the 2400 templates corresponds to a total of $512\times512=262144$ potential targets. The specific methods involved in this experiment are
\begin{itemize}
	\item	The geometric invariant representation SID \cite{ref81}, with FM-like robust design for rotation variants; 
	\item	The state-of-the-art dense texture representation DAISY \cite{ref82} and DASC \cite{ref83}, with robust design for geometric or signal variants;
	\item	The state-of-the-art deep learning representation DDIS \cite{ref84}, with robust design for nonlinear deformation.
\end{itemize}

\begin{figure*}[!t]
	\centering
	\includegraphics[scale=0.75]{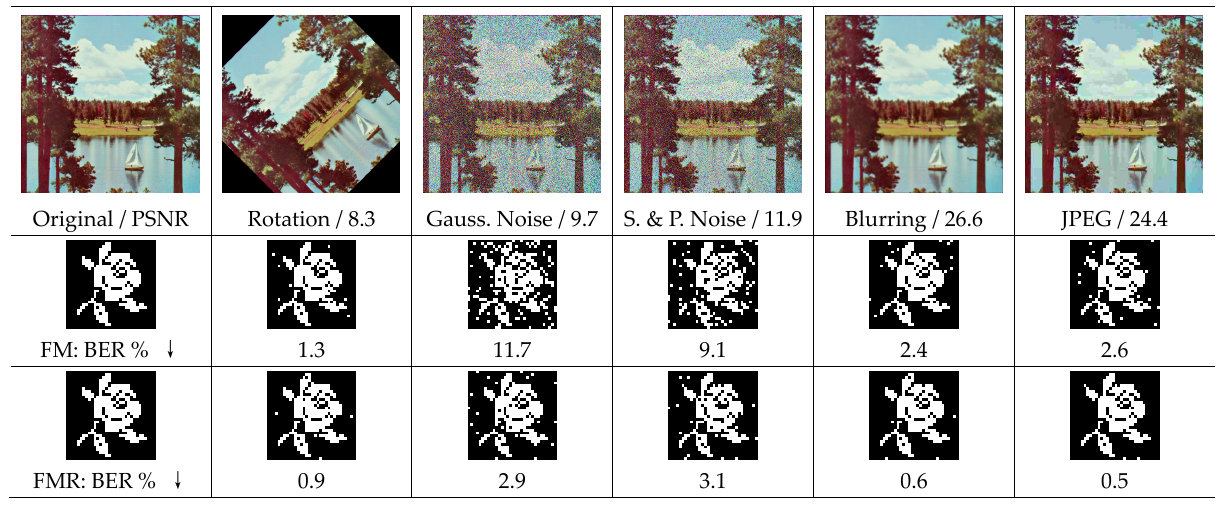}
	\centering
	\caption{Some degraded samples of the original image and the corresponding retrieved copyright codes from the zero-watermarking algorithms based on FM and FMR.}
\end{figure*}

In Tables 4 and 5, we present the accuracy scores for above representations on microscopy and smartphone images, respectively. Here, the top/bottom ranked performance is highlighted in red/blue.

One can observed that such methods show significant performance differences across the four experimental protocols, implying the difficulty of template matching in the wild. Due to sensitivity to imaging noise, rotation-invariant SID fails to achieve satisfactory matching accuracy, even for rotation-only protocol. Note that this phenomenon also illustrates the need to introduce FMR-like designs in the research community of invariants. In contrast, DAISY and DASC exhibit better performance in the ideal protocol due to higher discriminability. However, both lack in-form geometric invariance, leading to the failure in protocols containing rotations. The deep representation DDIS, designed specifically for template matching in the wild, achieves quite consistent accuracy performance across the four protocols. However, such robustness may come at the cost of discriminability, with a lower level of matching accuracy in general. This is mainly due to the fundamental difficulty of enabling invariance in deep representations. As for the proposed FMR, it exhibits significantly better and more stable performance w.r.t. two datasets and four protocols. This phenomenon suggests that FMR achieves the theoretically expected rotation invariance and noise robustness on real noisy images, with also sufficient discriminability for template matching tasks.

\begin{figure*}[!t]
	\centering
	\includegraphics[scale=0.89]{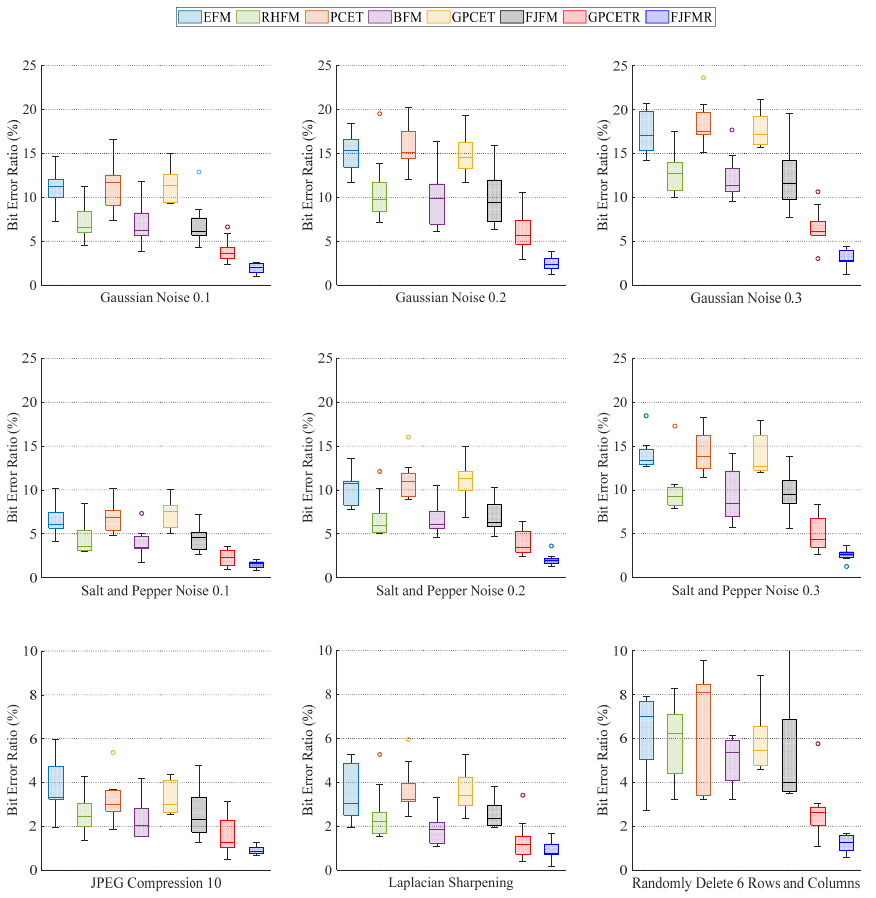}
	\centering
	\caption{ Box-plots of the bit error ratio by zero-watermarking algorithms with different representations in robustness experiment.}
\end{figure*}

\subsection{Zero-watermarking}

Unlike many classical vision tasks (e.g., image classification), visual security tasks (e.g., image forensics) are highly adversarial — they design more robust security tools, and then more powerful attacks by the adversary \cite{ref5}. In such two-player arms race, the representation robustness for image noise, e.g., photographic noise, transmission noise, and editing noise, is clearly the desired property.

In this paper, we consider the application of FMR in zero-watermarking for image copyright protection. The zero-watermarking is an alternative idea to traditional watermarking. Instead of embedding copyright in image, it records a binary sequence (i.e., zero-watermark) binding the image content and copyright in a trusted database (especially the blockchain) \cite{ref45}. Technically, such zero-watermark is the XOR of the content-aware code from original image (i.e., perceptual hash) and the copyright code (e.g., author’s ID). During the verification, the XOR of content-aware code from the testing image and the zero-watermark will recover the copyright code if such testing image is a duplication of the original one.

Here, a fundamental difficulty is ensuring that the content-preserving variants (e.g., noisy versions) of the original image are properly authenticated by the zero-watermarking. We attempt to improve the robustness of the zero-watermarking algorithm by replacing its representations with FMR, providing both rotation-invariant and noise-robust copyright protection. For a fair experimental comparison, the algorithms follow the same framework with difference only in image representation, hence better reflecting the inherent performance of such representation in zero-watermarking task. Furthermore, the representation vector is of the same dimension as the copyright code, and the representation parameters, e.g. $n,m,p,q,\alpha$ are generated randomly. 

In Fig. 12, we illustrate the robustness for the zero-watermarking algorithms w.r.t. different representations and attacks. The magnitudes of harmonic FMR and its FM counterpart are chosen here as examples of representation. Here, the original image is degraded to form a content-preserving variant, and then such variant is verified by zero-watermarking algorithm to retrieve the copyright code. Two widely known metrics, Peak Signal to Noise Ratio (PSNR) and Bit Error Ratio (BER), are introduced to measure the fidelity of image variant and the degradation of retrieved copyright code, respectively.

It can be observed that the copyright codes retrieved from the FMR-based algorithm exhibit consistently lower BER, especially under the severe noise where PSNR$\sim$10. This common advantage over the image-domain FM representations is in line with our theoretical expectations (Property 3) and is also consistent with the pattern recognition experiments (Section 5.3). Note that although both FM and FMR have perfect rotation invariance under the continuous-domain assumption, the resampling/quantization errors in discrete-domain rotation may affect this invariance in practice. 

In Fig. 13, we present the corresponding robustness box-plots with comprehensive representations and attacks, revealing the distribution properties of the BER scores. This experiment covers several major competing algorithms in the field of zero-watermarking with EFM \cite{ref59}, RHFM \cite{ref60}, PCET \cite{ref61}, BFM \cite{ref62}, GPCET \cite{ref45}, and FJFM \cite{ref23} representations. Here, our harmonic FMR and polynomial FMR based algorithms are also named as GPCETR and FJFMR in line with above notations and Section 6.3.

It is clear from Fig. 13 that, compared with such existing representations in the field of zero-watermarking, the proposed FMR versions (especially FJFMR) exhibit lower BER deviations while maintaining lower average BER. This phenomenon remains consistent for different attack types and intensities. Therefore, these BER statistics provide strong evidence for the uniqueness and usefulness of the proposed FMR in the zero-watermarking task.

\section{Conclusion}

The main goal of this paper is to provide a comprehensive design of robust representation for noisy images. We name this complete pipeline as ``Fractional-order Moments in Radon space (FMR)”, which is characterized by noise robustness, rotation invariance, and time-frequency discriminability.

The \emph{theory} ingredients of our work are as follows.
\begin{itemize}
	\item Regarding the definition, the implicit path (Section 4.1) and explicit path (Section 5) for constructing the FMR are both discussed in detail. The implicit definition is more intuitive and thus facilitates the derivation of key properties and the design of efficient implementations. The explicit definition reveals that the FMR can be interpreted as an infinite linear combination of geometric moments, providing mathematical material for future research.  
	\item Regarding the property, the generic nature, rotation invariance, noise robustness, and time-frequency discriminability are is analyzed in depth (Section 4.2). Such beneficial properties draw a distinction between the FMR and some theoretically relevant methods, implying the uniqueness and usefulness in practical visual tasks (Table 1).
	\item Regarding the implementation, the computationally efficient and numerically stable solutions are designed by Fourier transform and recursive formula (Section 4.3). Such implementations promote the practical applications of FMR based on discrete calculations. 
\end{itemize}

The \emph{practice} ingredients of our work are as follows.
\begin{itemize}
	\item Regarding the simulation experiment, the noise robustness and discriminability of FMR are first visualized by the histogram and reconstruction (Sections 6.1 and 6.2). The comprehensively benchmarking for FMR, image-domain counterpart, and learning representations (with augmentation/denoising) are then performed by a pattern recognition study on challenging noisy images (Sections 6.3). Such overall robustness statistics fully validated our main ideas w.r.t. rotation invariance, noise robustness, and time-frequency discriminability.
	\item Regarding the specific application, template matching in the wild (Sections 6.4) and zero-watermarking for copyright protection (Sections 6.5) are introduced due to their nature, i.e., robustness is desired. Such direct applications of FMR globally improve the robustness for various (signal and geometric) distortions, than popular representations of two tasks. In general, above state-of-the-art performance of FMR proves its promise in robust visual problems.
\end{itemize}

\ifCLASSOPTIONcaptionsoff
  \newpage
\fi



%

\bibliographystyle{IEEEtran}
\bibliography{paper}



%

\begin{IEEEbiography}[{\includegraphics[width=1in,height=1.25in,clip,keepaspectratio]{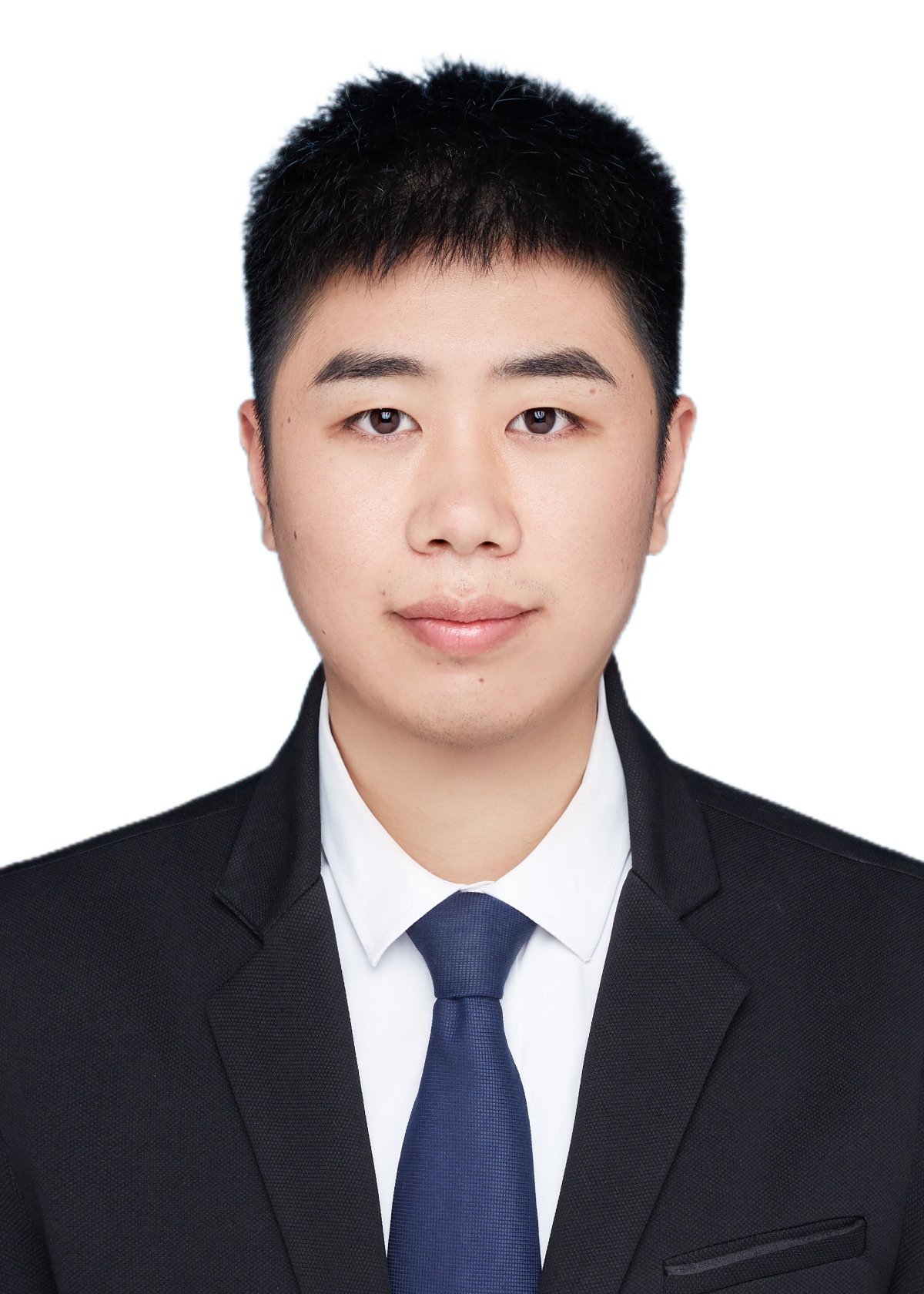}}]{Shuren Qi}
	received the Ph.D. degree in computer science from Nanjing University of Aeronautics and Astronautics, Nanjing, China, in 2024. He has published academic papers in top-tier venues including \emph{ACM Computing Surveys} and \emph{IEEE Transactions on Pattern Analysis and Machine Intelligence}. His research involves the general topics of invariance, robustness, and explainability in computer vision, with a focus on invariant representations, for closing today's trustworthiness gap in artificial intelligence, e.g., forensic and security of visual data.
\end{IEEEbiography}
\begin{IEEEbiography}[{\includegraphics[width=1in,height=1.25in,clip,keepaspectratio]{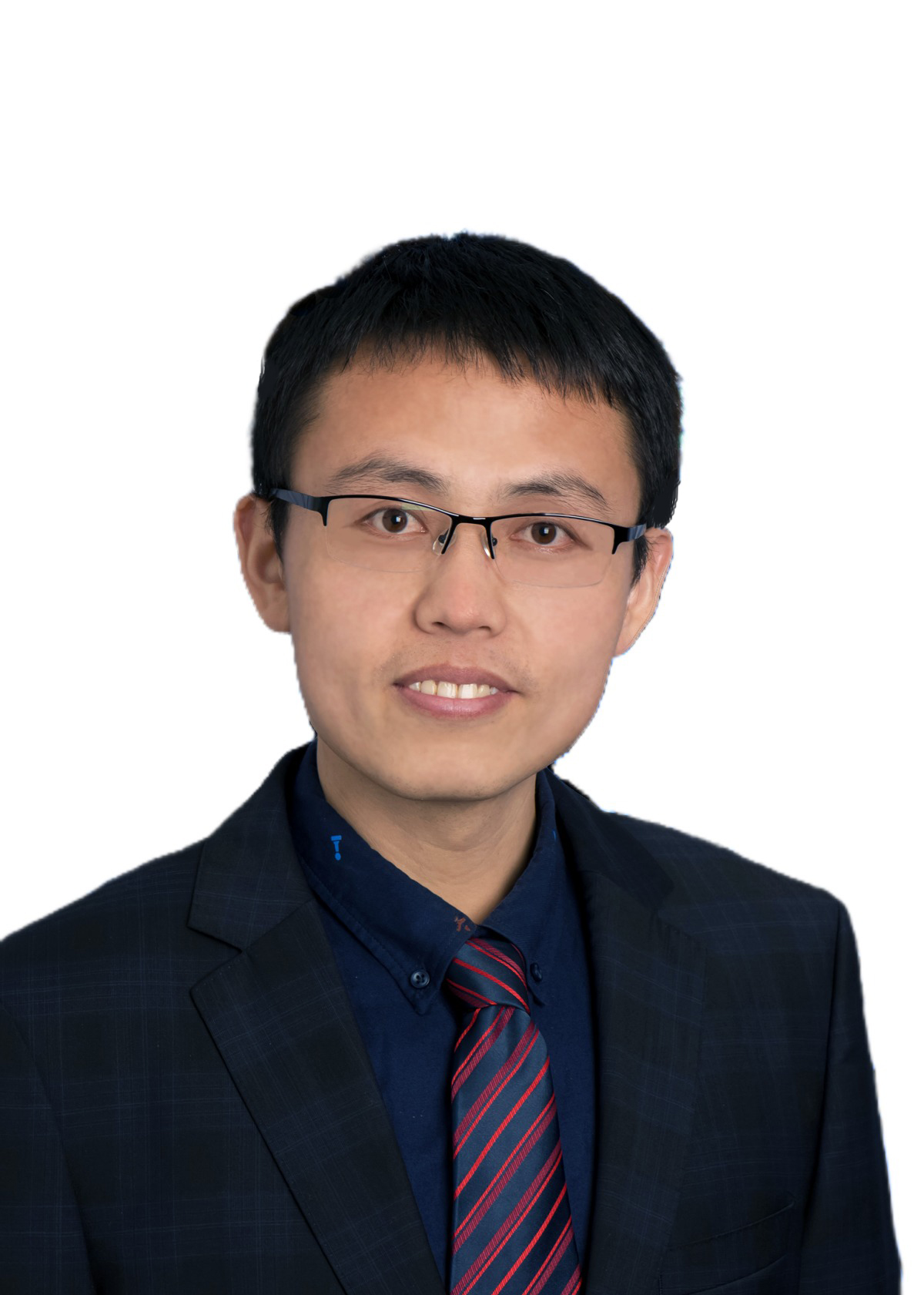}}]{Yushu Zhang}
	(Senior Member, IEEE) received the Ph.D. degree in computer science from Chongqing University, Chongqing, China, in 2014. He held various research positions with the City University of Hong Kong, Southwest University, University of Macau, Deakin University, and Nanjing University of Aeronautics and Astronautics. He is currently a Professor with the College of Information Technology, Jiangxi University of Finance and Economics, Nanchang, China. His research interests include multimedia processing and security, artificial intelligence, and blockchain. Dr. Zhang is an Associate Editor of \emph{Signal Processing} and \emph{Information Sciences}.
\end{IEEEbiography}

\begin{IEEEbiography}[{\includegraphics[width=1in,height=1.25in,clip,keepaspectratio]{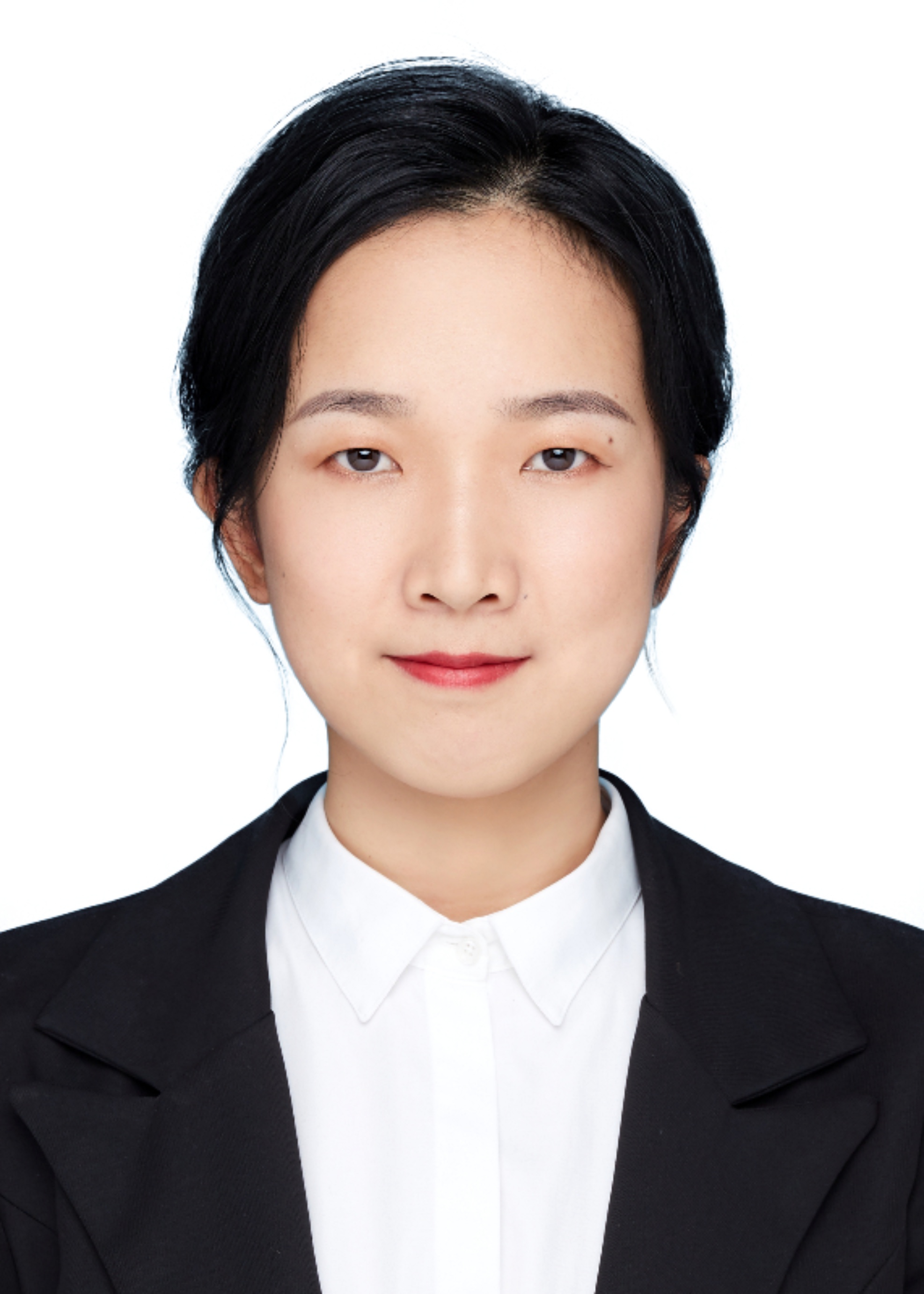}}]{Chao Wang}
	received the B.S. and M.S. degrees from Liaoning Normal University, Dalian, China, in 2017 and 2020 respectively. She is currently pursuing the Ph.D. degree in computer science at Nanjing University of Aeronautics and Astronautics, Nanjing, China. Her research interests include trustworthy artificial intelligence, adversarial learning, and media forensics.
\end{IEEEbiography}

\begin{IEEEbiography}[{\includegraphics[width=1in,height=1.25in,clip,keepaspectratio]{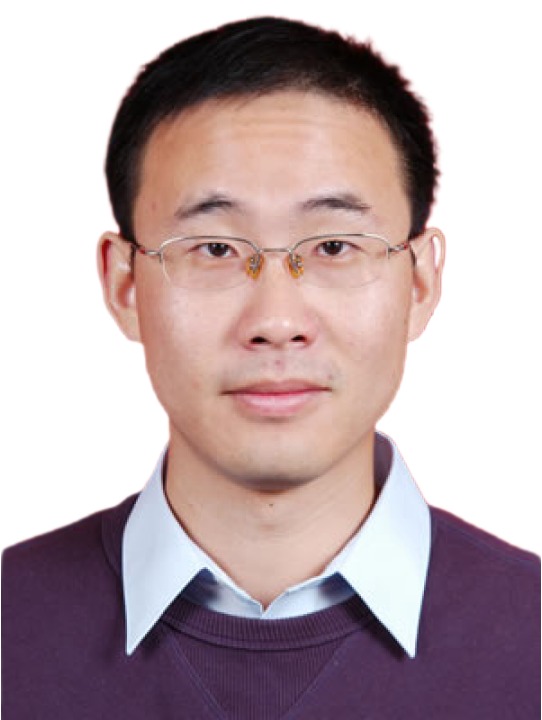}}]{Tao Xiang}
	(Senior Member, IEEE) received the B.E., M.S., and Ph.D. degrees in computer science from Chongqing University, China, in 2003, 2005, and 2008, respectively. He is currently a Professor with the College of Computer Science, Chongqing University, Chongqing, China. His research interests include multimedia security, cloud security, data privacy, and cryptography. He has published over 90 papers in international journals and conferences. He also served as a referee for numerous international journals and conferences.
\end{IEEEbiography}

\begin{IEEEbiography}[{\includegraphics[width=1in,height=1.25in,clip,keepaspectratio]{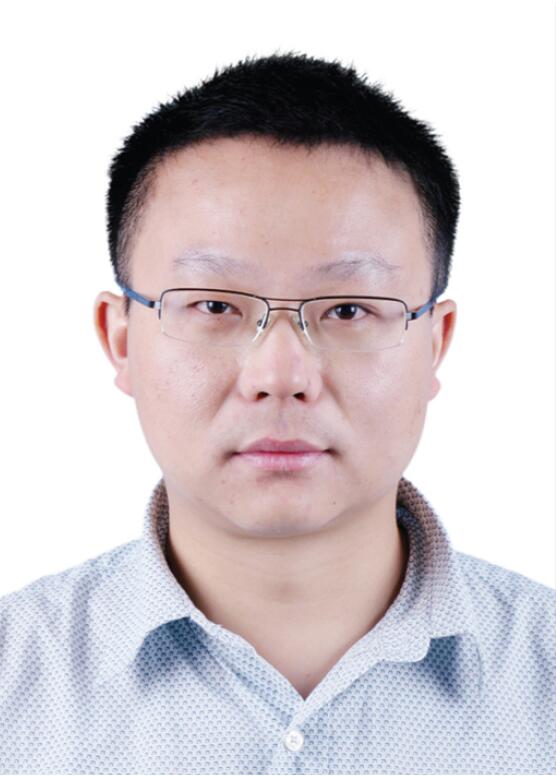}}]{Xiaochun Cao}
	(Senior Member, IEEE) received the B.E. and M.E. degrees in computer science from Beihang University, Beijing, China, in 1999 and 2002, respectively, and the Ph.D. degree in computer science from the University of Central Florida, Orlando, FL, USA, in 2006. After graduation, he spent about three years at ObjectVideo Inc., as a Research Scientist. From 2008 to 2012, he was a Professor at Tianjin University, Tianjin, China. Before joining Sun Yat-sen University, Shenzhen, China, he was a Professor at the Institute of Information Engineering, Chinese Academy of Sciences, Beijing, China. He is a Professor and the Dean with the School of Cyber Science and Technology, Shenzhen Campus of Sun Yat-sen University. He has published more than 200 journal and conference papers. Dr. Cao’s dissertation was nominated for the University Level Outstanding Dissertation Award. He was a recipient of the Piero Zamperoni Best Student Paper Award at the \emph{International Conference on Pattern Recognition}, in 2004 and 2010. He was on the Editorial Boards of \emph{IEEE Transactions on Circuits and Systems for Video Technology} and \emph{IEEE Transactions on Multimedia}. He is on the Editorial Boards of \emph{IEEE Transactions on Pattern Analysis and Machine Intelligence} and \emph{IEEE Transactions on Image Processing}.
\end{IEEEbiography}

\begin{IEEEbiography}[{\includegraphics[width=1in,height=1.25in,clip,keepaspectratio]{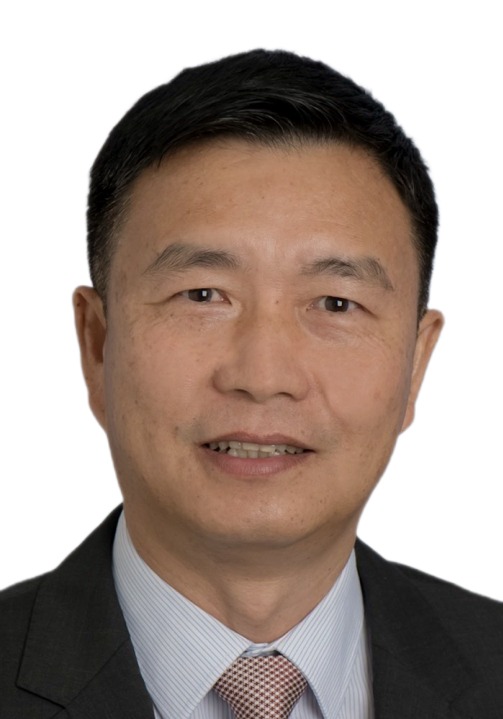}}]{Yong Xiang}
	(Senior Member, IEEE) received the Ph.D. degree in Electrical and Electronic Engineering from The University of Melbourne, Australia. He is a Professor at the School of Information Technology, Deakin
	University, Australia. His research interests include distributed computing, cybersecurity and privacy,
	machine learning and AI, and communications technologies. He has published 7 monographs, over 220 refereed journal articles, and over 100 conference papers in these areas. Prof. Xiang is the Senior
	Area Editor of \emph{IEEE Signal Processing Letters}, the Associate Editor of \emph{IEEE Communications Surveys and Tutorials}, and the Associate Editor of \emph{Computer Standards and Interfaces}. He was the Associate Editor of \emph{IEEE Signal Processing Letters} and \emph{IEEE Access}, and the Guest Editor of \emph{IEEE Transactions on Industrial Informatics}, \emph{IEEE Multimedia}, etc. He has served as Honorary Chair, General Chair, Program Chair, TPC Chair, Symposium Chair and Track Chair for many conferences, and was invited to give keynotes at numerous international conferences.
\end{IEEEbiography}




\end{document}